%% file: main.tex
\documentclass[10pt,twocolumn,letterpaper]{article}

\usepackage[accsupp]{axessibility}
\usepackage{cvpr}              
\usepackage{booktabs}
\usepackage{multirow}
\usepackage{adjustbox}
\usepackage{graphicx}
\usepackage{xcolor} 
\usepackage{makecell} 
\usepackage[table]{xcolor}
\usepackage{pdflscape} 
\usepackage{lscape}  
\usepackage{tcolorbox}
\usepackage{caption}
\usepackage{float}
\usepackage{relsize}
\usepackage{lineno}

\input{preamble}

\definecolor{cvprblue}{rgb}{0.21,0.49,0.74}
\usepackage[pagebackref,breaklinks,colorlinks,allcolors=cvprblue]{hyperref}

\def\paperID{} 
\def\confName{CVPR}
\def\confYear{2026}

\title{AnatomiX, an Anatomy-Aware Grounded Multimodal Large Language Model for Chest X-Ray Interpretation}

\author{Anees Ur Rehman Hashmi\\
Hasso Plattner Institute\\
Potsdam, Germany\\
{\tt\small anees.hashmi@hpi.de}
\and
Numan Saeed\\
MBZUAI\\
Abu Dhabi, UAE\\
{\tt\small numan.saeed@mbzuai.ac.ae}
\and
Christoph Lippert\\
Hasso Plattner Institute\\
Potsdam, Germany\\
{\tt\small christoph.lippert@hpi.de}
}

\begin{document}

\maketitle
\input{sec/0_abstract}    
\input{sec/1_intro}
\input{sec/related_work}
\input{sec/method}

\input{sec/experiments}

\input{sec/results}
\input{sec/conclusion}
{
    \small
    \bibliographystyle{ieeenat_fullname}
    \bibliography{main}
}

\input{sec/X_suppl}

\end{document}

%% file: preamble.tex
\definecolor{embcolor}{RGB}{0,120,215}   
\definecolor{boxcolor}{RGB}{200,60,0}    
\definecolor{bracketcolor}{RGB}{0,150,0} 
\newcommand{\embtag}[1]{\textcolor{embcolor}{#1}}
\newcommand{\boxtag}[1]{\textcolor{boxcolor}{#1}}
\newcommand{\brk}[1]{\textcolor{bracketcolor}{#1}}



%% file: sec/0_abstract.tex
\begin{abstract}

Multimodal medical large language models have shown substantial progress in chest X-ray interpretation but continue to face challenges in spatial reasoning and anatomical understanding. Although existing grounding techniques improve overall performance, they often fail to establish a true anatomical correspondence, resulting in incorrect anatomical understanding in the medical domain. To address this gap, we introduce AnatomiX, a multitask multimodal large language model for anatomically grounded chest X-ray interpretation. Inspired by the radiological workflow, AnatomiX adopts a two stage approach: first, it identifies anatomical structures and extracts their features, and then leverages a large language model to perform diverse downstream tasks such as phrase grounding, report generation, visual question answering, and image understanding. Extensive experiments across multiple benchmarks demonstrate that AnatomiX achieves superior anatomical reasoning and delivers over 25\% improvement in performance on anatomy grounding, phrase grounding, grounded diagnosis and grounded captioning tasks compared to existing approaches. Code and pretrained model are available at \href{https://aneesurhashmi.github.io/anatomix/}{aneesurhashmi.github.io/anatomix}.


\end{abstract}

%% file: sec/1_intro.tex
\begin{figure*}[t]
  \centering
  \hfill
  \includegraphics[width=\linewidth]{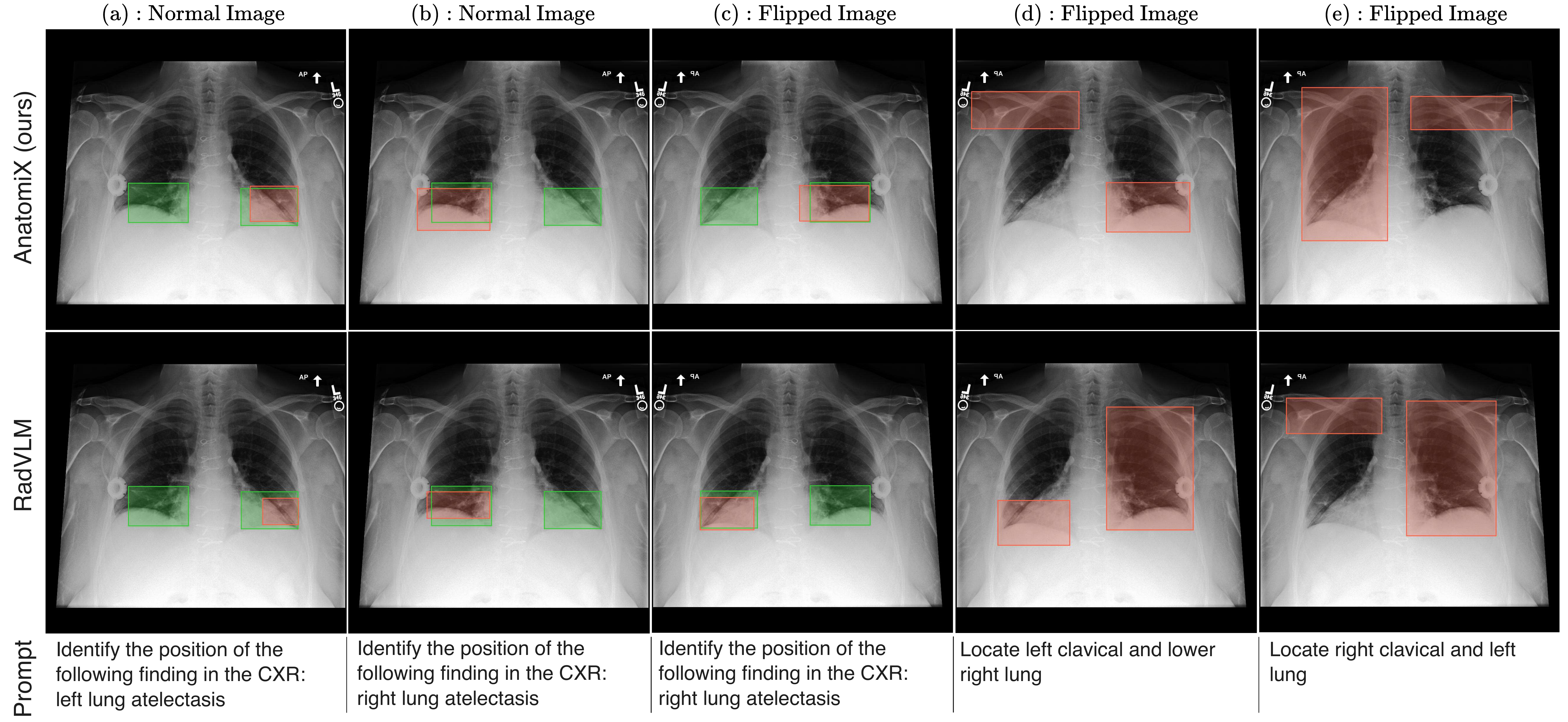}
  \caption{Comparison between AnatomiX and RadVLM \cite{deperrois2025radvlm} in anatomy understanding. (a) and (b) show both models predicting the disease on the correct side (color scheme: red for model's output, green for all ground truth locations). (c), (d) and (e) show models' outputs for the same image flipped on the vertical axis (left $\leftrightarrow$ right), where RadVLM completely fails to recognize the correct anatomical object, while AnatomiX successfully recognizes the correct anatomies, showcasing high anatomical understanding} \label{fig:fig_left_right}
\end{figure*}
\section{Introduction} \label{sec:intro}

Multimodal Large Language Models (MLLMs) are being increasingly applied in the natural and medical imaging domain to perform multiple tasks using a single model \cite{xiao2024comprehensive}. These models typically consist of an image encoder and a Large Language Model (LLM), and utilize the pretrained LLM’s strengths by passing image embeddings along with a text prompt into the LLM to perform downstream tasks \cite{liu2023visual}. The pretrained LLMs are generally trained on very large text corpora and therefore demonstrate strong text generation capabilities, making them suitable for a diverse set of downstream tasks after supervised fine-tuning and instruction tuning \cite{anisuzzaman2024fine, ma2024groma}. However, owing partly to their autoregressive design and the challenges of merging vision and language modalities, MLLMs still struggle with fine-grained spatial understanding, for instance, when reasoning about positions of multiple objects or their relative spatial relations in a scene \cite{liu2025can, wolf2025your}. 

This issue of MLLMs has previously been addressed by introducing object grounding, which aligns the text concepts with the objects in the image \cite{szot2024grounding, peng2023kosmos}. Grounding in MLLMs is usually achieved by training the model on a dataset containing the names or descriptions of local objects in the image plus their bounding boxes or segmentation masks as spatial markers. This in turn improves the reasoning abilities of MLLMs with better concept understanding, making them applicable in the medical domain, where spatial reasoning is essential. In particular, chest X-ray (CXR) interpretation greatly benefits from such multimodal reasoning, as accurate localization and semantic alignment between textual findings and radiographic regions are crucial for diagnosis. MLLMs like ViviMed \cite{luo2024vividmed}, ChexAgent \cite{chen2024chexagent}, RadVLM \cite{deperrois2025radvlm} and MAIRA-2 \cite{bannur2024maira} show that grounding via special tokens yields consistent performance gains on CXR image tasks.



Although incorporating grounding through additional tokens has improved MLLMs’ spatial reasoning, it remains insufficient for the fine-grained localization and differentiation required in medical imaging, where anatomically distinct regions often exhibit highly similar visual textures and appearances \cite{wolf2025your}. As illustrated in Fig. \ref{fig:fig_left_right}, current state-of-the-art (SOTA) MLLM fails to correctly localize lesions or identify the correct anatomical objects when presented with  flipped images - where the left and right sides are switched. These models may perform well on standard orientations but fail when spatial cues are inverted, revealing that they overly rely on spatial correlations rather than recognition of anatomical structures, exposing a critical gap between visual grounding and medical comprehension.

This weak anatomical understanding in current medical MLLMs is likely due to their single-step visual grounding process. Specifically, these models must implicitly detect the correct anatomical objects within an image before performing the downstream task. This one-step process differs fundamentally from the workflow of radiologists, who iteratively identify, localize, and evaluate each anatomical structure before drawing diagnostic conclusions. To address this issue, we introduce \textit{AnatomiX}, an anatomy-aware grounded MLLM for chest X-Ray interpretation. AnatomiX uses a two stage process to first identify different thoracic anatomical objects (organs) before performing the task; thereby, showing a high anatomical understanding compared to existing CXR grounding MLLMs as shown in Fig. \ref{fig:fig_left_right}. Our proposed model significantly outperforms SOTA models on four grounding tasks and shows SOTA or on-par performance on report generation, VQA and image understanding tasks. Extensive experiments on a large collection of datasets show the high reasoning and anatomical understanding capabilities of AnatomiX. In summary, our work makes the following contributions:

\begin{itemize}
    \item We introduce \textit{AnatomiX}, an anatomy-aware grounded multimodal large language model for CXR interpretation.
    \item AnatomiX improves anatomical understanding by achieving SOTA performance on diverse grounding tasks, while maintaining on-par or better performance on report-generation, VQA and image understanding tasks.
    \item We demonstrate the robustness of AnatomiX across different datasets and challenging settings, and validate the contribution of each component in ablation experiments.
    
\end{itemize}







%% file: sec/related_work.tex
\section{Related Work}
Early adaptations of MLLMs to radiology primarily involve fine-tuning general-domain models on medical datasets. LLaVA-Med \cite{li2023llava} and RadVLM \cite{deperrois2025radvlm} extend LLaVA \cite{liu2023visual} for multi-task CXR benchmarks, improving both report generation and VQA. Several works introduce explicit grounding for CXR tasks: ViviMed \cite{luo2024vividmed} and MedRG \cite{zou2024medrg} pair the Segment-Anything model \cite{kirillov2023segment} with an LLM for detection and segmentation, while MAIRA-2 \cite{bannur2024maira} enables grounded report generation through additional tokens. Similarly, RadVLM \cite{deperrois2025radvlm} constructs a large instruction dataset for diverse CXR tasks, whereas CheXagent \cite{chen2024chexagent} applies contrastive learning and instruction tuning to enhance phrase grounding and CXR report generation. Radialog \cite{pellegriniradialog} supports multi-turn CXR conversations, and MedGemma \cite{sellergren2025medgemma} adapts Gemma-3 for general purpose medical tasks by finetuning on large-scale medical datasets. More recently, AOR \cite{li2025aor} introduced region level information in LLM for CXR interpretation; however, this model is not yet publicly available for testing and comparison.

\begin{figure*}[!t]
  \centering
  \hfill
  \includegraphics[width=\linewidth]{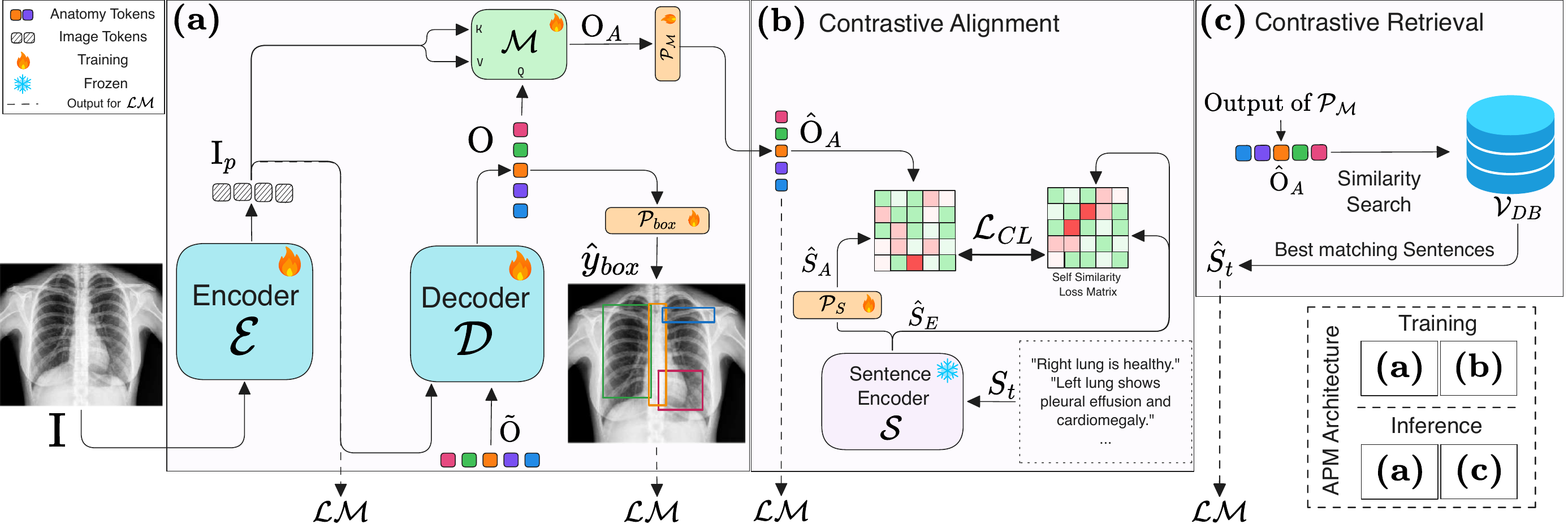}
  \caption{Anatomy Perception Module (APM) architecture $\mathrm{\textbf{(a)}}$: The encoder $\mathcal{E}$ outputs image embedding $\mathrm{I}_p$, while the decoder $\mathcal{D}$ and feature extraction module $\mathcal{M}$ output object bounding boxes $\hat{y}_{box}$, and anatomical object tokens $\hat{\mathrm{O}}_A$, respectively. Different colors in $\Tilde{\mathrm{O}}$, $\mathrm{O}$ and $\hat{\mathrm{O}}_A$ represent specific anatomical objects. $\mathrm{\textbf{(b)}}$ shows the contrastive alignment using frozen sentence encoder $\mathcal{S}$ and self-similarity loss. $\mathrm{\textbf{(c)}}$: The vector database ($\mathcal{V}_{DB}$) contains the text sentences and embeddings used for contrastive retrieval. (Bottom right): APM uses $\mathrm{\textbf{(a)}}$ and $\mathrm{\textbf{(b)}}$ during training, and replaces $\mathrm{\textbf{(b)}}$ with $\mathrm{\textbf{(c)}}$ during inference. The $\mathcal{P}$ represent different FC projectors described in section \ref{sec:stage_1}.}
  \label{fig:fig_stage1}
\end{figure*}

Most prior efforts adapt general-domain MLLMs rather than developing domain-specific architectures. CheX \cite{muller2024chex} advances toward anatomy-aware modeling by incorporating anatomical objects from CXR reports but lacks prompt-based interaction and generative flexibility. Overall, existing CXR MLLMs rely on instruction tuning and large-scale radiology datasets, achieving strong benchmark performance yet showing limited reasoning and anatomical understanding \cite{wolf2025your}. In contrast, our approach introduces a two-stage anatomy-aware pipeline that explicitly models thoracic structures before performing downstream tasks.

%% file: sec/method.tex
\section{Methodology} \label{sec:method}

This section describes the architecture of AnatomiX, which comprises two primary components: The Anatomy Perception Module and a large language model outlined below.

\subsection{Anatomy Perception Module} \label{sec:stage_1}

Given an input CXR image $\mathrm{I}$, the objective of the Anatomy Perception Module (APM) is to extract a global image representation along with fine-grained features corresponding to $N$ thoracic anatomical objects. The resulting representations are subsequently used by an LLM for downstream tasks. The APM adopts a multi-task learning framework that jointly learns global image features, object localization through bounding boxes, and their detailed anatomical representations, while also retrieving textual descriptions associated with each anatomical object. Fig. \ref{fig:fig_stage1} shows the detailed architecture of the APM, which consists of an image encoder ($\mathcal{E}$), a decoder ($\mathcal{D}$), a feature extraction module ($\mathcal{M}$) and a sentence encoder ($\mathcal{S}$). The details of each component are given below. \\

\noindent \textbf{Image Encoder $\mathcal{E}$ and Decoder $\mathcal{D}$:} 

\noindent The input image $\mathrm{I}$ is first encoded by the image encoder $\mathcal{E}$ producing the image representation $\mathrm{I}_p \in \mathbb{R}^{P \times d}$ which consists of $P$ patch embedding vectors of dimension $d$. Together, these form a global representation of the image. These embeddings serve as the shared representation of visual information for subsequent modules. Specifically, $\mathrm{I}_p$ is provided both to the feature extractor $\mathcal{M}$, which focuses on semantic anatomy cues, and to the decoder $\mathcal{D}$, which is inspired by DETR \cite{carion2020end}. The decoder processes $\mathrm{I}_p$ jointly with $N$ learnable object tokens $\Tilde{\mathrm{O}} = [\Tilde{o}^1, \ldots, \Tilde{o}^N]$, using transformer blocks to perform cross-attention between tokens and image patches. Through this interaction, each object token learns to attend to relevant anatomical regions, resulting in updated token embeddings $\mathrm{O}$ that encode the localization of the $N$ anatomical objects.

\begin{equation}
    \mathrm{O} = \mathcal{D}(\mathrm{I}_p,\Tilde{\mathrm{O}}) 
    \label{eq:image_decoder}
\end{equation}

\noindent where $\mathrm{O} \in \mathbb{R}^{N \times d} = [o^1, ..., o^N; o^i \in \mathbb{R}^d ]$

The object tokens $\mathrm{O}$ are projected using a fully connected (FC) projector $\mathcal{P}_{box}$ to predict a bounding box for each anatomical object ($\hat{y}_{box} = \mathcal{P}_{box}(\mathrm{O})$), as illustrated in Fig. \ref{fig:fig_stage1}. Unlike DETR \cite{carion2020end}, $\mathrm{O}$ is not permutation invariant, which means that each element of $\mathrm{O}$ corresponds to exactly one predefined anatomical object. This design enables each token $o^{i}$ to focus on and extract information related to the $i^{th}$ anatomical object from $\mathrm{I}_p$. To effectively learn the $\hat{y}_{box}$ localization, we use a combination of L1 and intersection over union (IoU) losses as shown in eq. \ref{eq:iou_loss} and \ref{eq:bbox_loss}.

\begin{equation}
    \mathcal{L}_{IoU}(\hat{y}_{box}, y_{box}) =  1 - \frac{|\hat{y}_{box} \cap y_{box}|}{|\hat{y}_{box} \cup y_{box}|}
    \label{eq:iou_loss}
\end{equation}

\begin{equation}
    \mathcal{L}_{box} = \lambda_1 (|\hat{y}_{box}-y_{box}|) +  \lambda_2 \mathcal{L}_{IoU}(\hat{y}_{box}, y_{box})
    \label{eq:bbox_loss}
\end{equation}

\noindent where $y_{box}$ represents ground truth bounded boxes, $\lambda_1 = 5$ and $\lambda_2 = 2$ are the weightings of the $\mathrm{L1}$ Loss and the $\mathcal{L}_{IoU}$ Loss, respectively, set to the default values used in DETR.\\

\begin{figure}
  \centering
  \hfill
  \includegraphics[width=\linewidth]{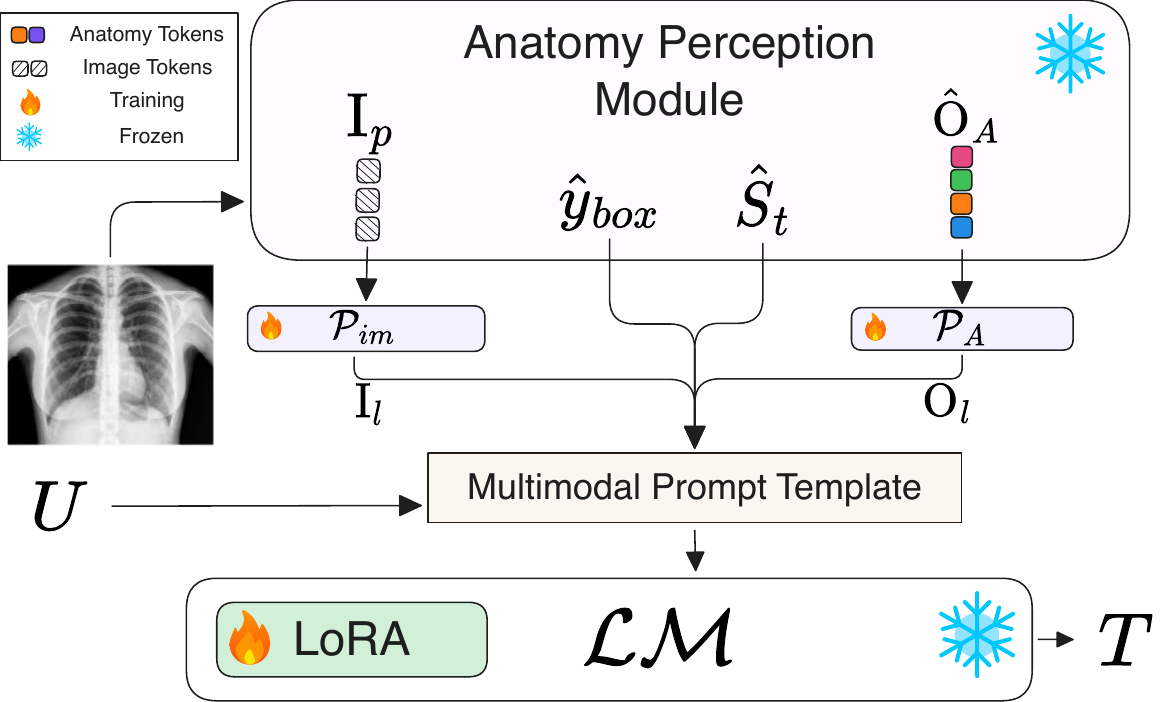}
  \caption{Overall architecture of AnatomiX. The outputs of the APM and the user prompt $U$ are added to a structured multimodal prompt template before being passed to the $\mathcal{LM},$ which generates the response $T$. $\mathcal{P}_{im}$ and $\mathcal{P}_{A}$ represent FC projectors as described in section \ref{sec:sec_llm}.}
  \label{fig:fig_stage2}
\end{figure}

\noindent \textbf{Feature Extraction Module $\mathcal{M}$:}  

\noindent In addition to predictions of the bounding boxes in the decoder output, we leverage the spatial information encoded in $\mathrm{O}$ to extract fine-grained representations of each anatomical object through the feature extraction module $\mathcal{M}$. Within $\mathcal{M}$, cross-attention is computed between $\mathrm{O}$ and image patches $\mathrm{I}_p$ as shown in eq. \ref{eq:anatomy_module}, where the anatomical object tokens $\mathrm{O}$ serve as queries ($Q$), and the image embedding $\mathrm{I}_p$ provides the keys ($K$) and values ($V$). Conceptually, $\mathcal{M}$ can be viewed as an extension of the decoder $\mathcal{D}$, where the image representation $\mathrm{I}_p$ is re-included via a skip connection.

\begin{equation}
    \mathrm{O}_A = \mathcal{M}(Q, K, V) = Softmax \left(\frac{Q K^T}{\sqrt{d}}\right) V
    \label{eq:anatomy_module}
\end{equation}

\noindent where $Q=\mathrm{O}\mathrm{W}_Q$ is the query matrix, $K=\mathrm{I}_p\mathrm{W}_K$ is the key matrix and $V=\mathrm{I}_p\mathrm{W}_V$ is the value matrix. \\

The output $\mathrm{O}_A$ represent localized features for the anatomical objects at the corresponding positions specified by the predicted bounding boxes $\hat{y}_{box}$ (see supp. Fig. A5). Subsequently, these features are projected to a lower-dimensional space ($d \rightarrow s$) using the projection module $\mathcal{P}_{\mathcal{M}}$ as: $\hat{\mathrm{O}}_A = \mathcal{P}_{\mathcal{M}}(\mathrm{O}_A)$, where $\hat{\mathrm{O}}_A \in \mathbb{R}^{N \times s}$. \\

\noindent \textbf{Contrastive Alignment with $\mathcal{S}$:} 

\noindent During APM training, we perform localized contrastive alignment (Fig. \ref{fig:fig_stage1}-b) between each anatomical feature token $\hat{o}_A^i \in \hat{\mathrm{O}}_A$ and its corresponding textual description $s^i \in S_t$, where $s^i$ specifies the radiological findings in the associated anatomical region (e.g. \textit{Right Lung shows pneumonia, pleural effusion and atelectasis}.). Since $\hat{\mathrm{O}}_A$ encodes fine-grained visual representations of $N$ anatomical objects derived from the image embedding $\mathrm{I}_p$, aligning these tokens with the corresponding text sentence embeddings enables the model to establish correspondences between visual and semantic representations. This alignment ensures that the visual tokens capture each anatomical region’s spatial and structural properties while linking them to clinically relevant textual concepts. The textual descriptions $S_t$ are first encoded using the frozen sentence encoder $\mathcal{S}$ (BiomedBERT \cite{gu2021domain}) to obtain embeddings $\hat{\mathrm{S}}_E \in \mathbb{R}^{N \times 768}$ as $\hat{\mathrm{S}}_E =\mathcal{S}(S_t)$. Finally, $\hat{\mathrm{S}}_E$ is projected to a lower-dimensional space ($768 \rightarrow s$) using a fully connected projector $\mathcal{P}_{S}$ for efficient contrastive learning as $\hat{\mathrm{S}}_A = \mathcal{P}_{S}(\hat{\mathrm{S}}_E)$, where $\hat{\mathrm{S}}_A  \in \mathbb{R}^{N \times s}$ represents projected text embeddings.



Radiological findings across thoracic anatomical objects often overlap or appear together, making it uncommon for an image to contain a finding in only one region. In such cases, using standard CLIP-style contrastive loss \cite{radford2021learning} can introduce many false negatives, as it assumes only one correct (positive) text–image pair per sample. To overcome this, we employ a soft contrastive loss that allows multiple degrees of similarity across anatomical regions. Specifically, we introduce a self-similarity matrix $\mathrm{S}_{self}$ (eq. \ref{eq:self_sim_matrix}), which permits non-zero similarity values for off-diagonal entries, reflecting the natural co-occurrence of anatomical observations (see the supp. material sec. S1 for details). We optimize this alignment using Kullback–Leibler (KL) divergence as shown in eq. \ref{eq:cl_loss}, which measures distributional differences instead of enforcing discrete class boundaries \cite{jiang2023cross}. This makes it well-suited for overlapping or correlated anatomical feature representations and preserving partial similarities among different anatomical objects' features. 




\begin{equation}
    \mathrm{S}_{self} =  \hat{\mathrm{S}}_E\hat{\mathrm{S}}_E^T
    \label{eq:self_sim_matrix}
\end{equation}
\begin{equation}
    \mathcal{L}_{CL}(\hat{\mathrm{S}}_A, \mathrm{S}_{self}) = \sum_{i} \hat{\mathrm{S}}_A(i) \log \frac{\hat{\mathrm{S}}_A(i)}{\mathrm{S}_{self}(i)}
    \label{eq:cl_loss}
\end{equation}

\noindent where $\mathrm{S}_{self} \in \mathbb{R}^{N \times N}$ \\


Finally, we train APM end-to-end using a combination of bounding box prediction and contrastive alignment losses.

\begin{equation}
    \mathcal{L}_{APM} =\mathcal{L}_{box} + \mathcal{L}_{CL}
    \label{eq:apm_loss}
\end{equation}

\noindent \textbf{Contrastive Retrieval with $\mathcal{V}_{DB}$:} 

\noindent Since the textual descriptions $S_t$ associated with anatomical objects are available only during APM training, we replace the sentence encoder $\mathcal{S}$ with a compact vector database $\mathcal{V}_{DB}$ during inference as shown in Fig. \ref{fig:fig_stage1}-(c). This database stores all unique textual sentences corresponding to each of the $N$ anatomical regions, along with their precomputed embeddings. At inference time, each anatomical object token in $\hat{\mathrm{O}}_{A}$ is compared against the sentence embeddings in $\mathcal{V}_{DB}$ to retrieve the most semantically similar sentence $\hat{S_t}$ for that anatomical region as $\hat{S}_t = \mathcal{V}_{DB}(\hat{\mathrm{O}}_A)$. The retrieved sentences are then passed to LLM, along with the anatomical object tokens $\hat{\mathrm{O}}_{A}$, the predicted bounding boxes $\hat{y}_{box}$, and the image embeddings $\mathrm{I}_p$. Additional implementation details of $\mathcal{V}_{DB}$ are provided in the supp. material sec. S2.



\subsection{Large Language Model} \label{sec:sec_llm}

Given the outputs of APM — image embeddings $\mathrm{I}_p$, predicted locations of $N$ anatomical objects $\hat{y}_{box}$ and their corresponding features $\hat{\mathrm{O}}_A$ and retrieved text descriptions $\hat{S}_t$ - and the user prompt $U$, the large language model $\mathcal{LM}$ generates a textual response $\mathrm{T}$ performing the task specified in $U$. To achieve this, we construct a multimodal prompt template (shown in supp. Fig. S2) that integrates the fine grained anatomical and textual information extracted in APM, and passes it to $\mathcal{LM}$. Fig. \ref{fig:fig_stage2} shows the overall architecture of AnatomiX.
Firstly, to align the embedding spaces of APM and $\mathcal{LM}$, we project both the image embedding $\mathrm{I}_p$ and anatomical object tokens embedding $\hat{\mathrm{O}}_A$ into $\mathcal{LM}$’s embedding space using FC projectors. Specifically, $\mathcal{P}_{im}$ ($d \rightarrow l$) maps the image embedding to $\mathrm{I}_{l} = \mathcal{P}_{im}(\mathrm{I}_p)$, where $\mathrm{I}_{l} \in \mathbb{R}^{P \times l}$, and $\mathcal{P}_{A}$ ($s \rightarrow l$) maps the anatomical object token embedding to $\mathrm{O}_{l} = \mathcal{P}_{A}(\hat{\mathrm{O}}_A)$, where $\mathrm{O}_{l} \in \mathbb{R}^{N \times l}$.






The language model $\mathcal{LM}$ is based on MedGemma-4b-it \cite{sellergren2025medgemma} LLM architecture (excluding vision encoder). To enable anatomy aware reasoning, we extend the model’s vocabulary by introducing $N$ special tokens for anatomical objects (\texttt{<obj\_i>} for $i \in [0, N]$) and four additional tokens for spatial grounding (\texttt{<box>}, \texttt{</box>}, \texttt{<ref>}, and \texttt{</ref>}). Each \texttt{<obj\_i>} token corresponds to the feature representation of the $i^{\text{th}}$ anatomical object, $\mathrm{o}_{l}^{i}$, where $\mathrm{O}_{l} = [\mathrm{o}_{l}^{1}, \dots, \mathrm{o}_{l}^{N}]$. By explicitly providing these object-specific tokens, the LLM can directly access the fine grained visual features of each anatomical object. This design allows $\mathcal{LM}$ to directly reason over anatomical objects rather than implicitly inferring them from global image representations and then performing the task given in $U$. Finally, the $\mathcal{LM}$ is trained for the next token prediction using Low-Rank Adaptation (LoRA) \cite{hu2022lora} and optimized with standard cross-entropy loss as shown in eq. \ref{eq:llm_loss}.

\begin{equation}
  \mathcal{L}_{\mathcal{LM}} =
  -\frac{1}{T}
   \sum_{t=1}^{T} \sum_{w \in V}
   y_t(w)\, \log p_{\mathcal{LM}}(w \mid x_{<t})
  \label{eq:llm_loss}
\end{equation}

\begin{equation}
  p_{\mathcal{LM}}(w \mid x_{<t})
  = \mathrm{Softmax}\!\left(\mathcal{LM}(x_{<t})\right)
  \label{eq:llm_softmax}
\end{equation}

\noindent where $T$ is the total number of tokens in the sequence, $V$ is the vocabulary, $x_{<t}$ represents the context tokens before position $t$, $y_t(w)$ is the one-hot ground truth distribution at step $t$, $w$ denotes a token in the vocabulary $V$, and $p_{\mathcal{LM}}(w \mid x_{<t})$ is the probability assigned by $\mathcal{LM}$ for token $w$ given $x_{<t}$.


%% file: sec/experiments.tex

\begin{figure}[!t]
  \centering
  \hfill
  \includegraphics[width=0.9\linewidth]{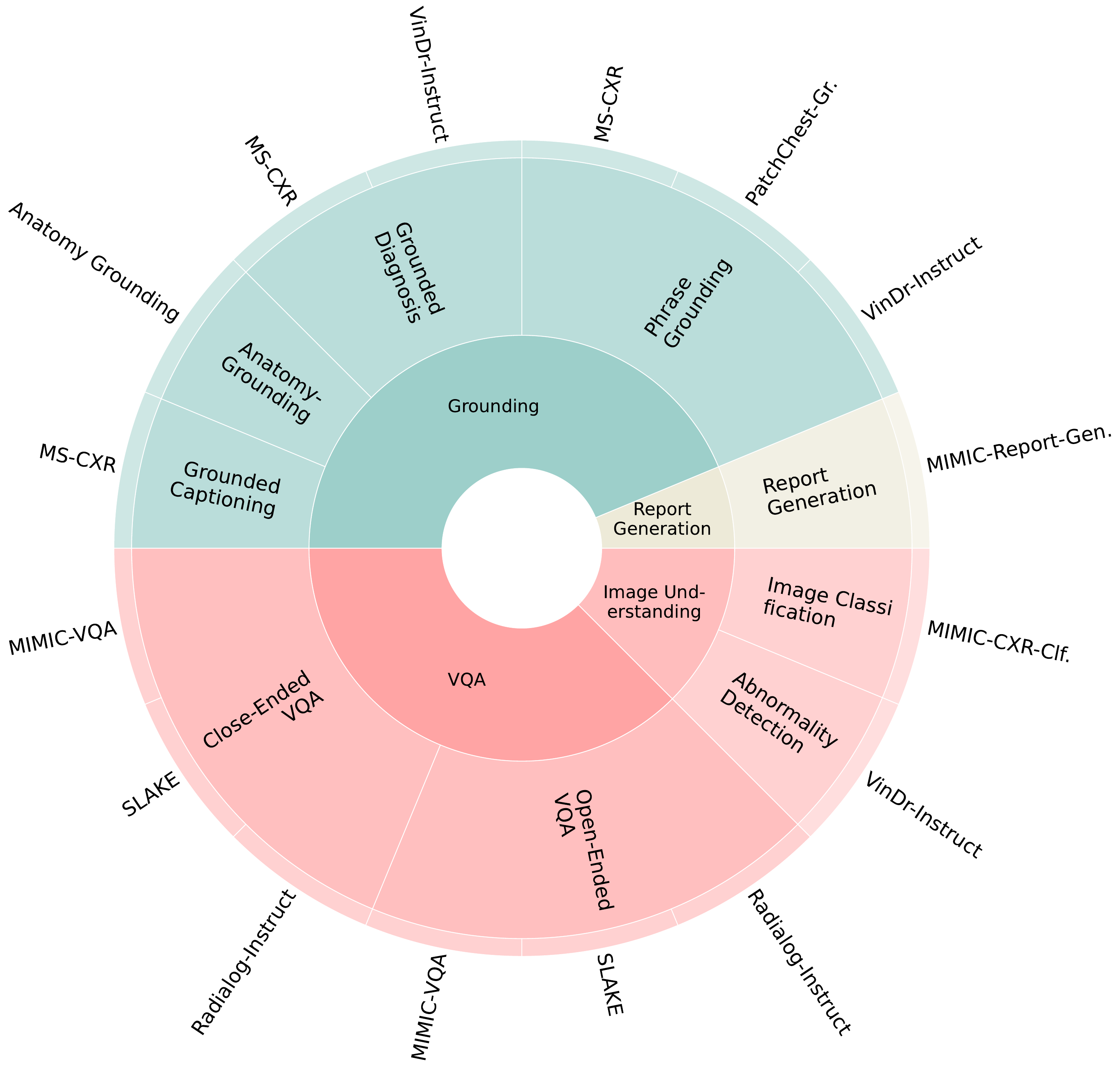}
  \caption{Set of 9 radiology tasks (middle circle) spanning 4 categories (inner circle) done by AnatomiX and the datasets used (outer circle).}
  \label{fig:fig_dataset_tasks}
\end{figure}

\setlength{\tabcolsep}{3.5pt} 
\begin{table*}[!t]
\caption{
Performance on four grounding tasks. 
For Grounded Diagnosis (GD) and Grounded Captioning (GC), results are shown as GD / GC. }
\centering
\scriptsize
\adjustbox{width=\textwidth}{
\begin{tabular}{
l|
*{3}{>{\centering\arraybackslash}p{0.075\textwidth}}|
*{2}{>{\centering\arraybackslash}p{0.075\textwidth}}|
*{2}{>{\centering\arraybackslash}p{0.07\textwidth}}|
*{2}{>{\centering\arraybackslash}p{0.07\textwidth}}
}
\toprule
\textbf{Model} &
\multicolumn{3}{c|}{\textbf{NLG Metrics (GD / GC)}} &
\multicolumn{2}{c|}{\textbf{Clinical Metrics (GD / GC)}} &
\multicolumn{2}{c|}{\textbf{Phrase Grounding}} &
\multicolumn{2}{c}{\textbf{Anatomy Grounding}} \\ 
\cmidrule(lr){2-4}\cmidrule(lr){5-6}\cmidrule(lr){7-8}\cmidrule(lr){9-10}
 & BERTScore & ROUGE & METEOR &
 RadGraph-F1  & Chexbert-14-F1 &
 IoU & mAP &
 IoU & mAP \\ 

\midrule

MAIRA-2   & 0.01 / 0.08 & 0.01 / 0.06 & 0.01 / 0.04 & 0.00 / 0.02 & 0.03 / 0.02 & 0.32 & 0.24 & 0.35 & 0.24 \\

RadVLM    & 0.15 / 0.27 & 0.06 / 0.11 & 0.05 / 0.07 & 0.00 / 0.12 & 0.32 / 0.40 & \underline{0.39} & \underline{0.30} & \underline{0.60} & \underline{0.49} \\
CheXagent & \underline{0.49} / \underline{0.56} & \underline{0.43} / \underline{0.44} & \underline{0.29} / \underline{0.37} & \underline{0.40} / \underline{0.39} & \underline{0.40} / \underline{0.61} & 0.33 & 0.24 & 0.18 & 0.09 \\
\textbf{AnatomiX (ours)} & \textbf{0.63} / \textbf{0.65} & \textbf{0.60} / \textbf{0.56} & \textbf{0.42} / \textbf{0.48} & \textbf{0.58} / \textbf{0.50} & \textbf{0.54} / \textbf{0.78} & \textbf{0.46} & \textbf{0.35} & \textbf{0.73} & \textbf{0.66} \\

\bottomrule

\end{tabular}
}
\label{tab:grounding-tasks}
\end{table*}

\section{Experiments}
\noindent \textbf{Dataset:} We train APM using over 237,000 samples from the Chest ImaGenome \cite{wu2021chest} dataset, which extends MIMIC-CXR \cite{johnson2019mimic} with detailed spatial and semantic annotations, providing localized information for 36 anatomical objects and their observations. We used the given object-wise phrases for $S_t$ and the given object bounding boxes for $y_{box}$ prediction.

Supp. Table S1 provides the summary of nine datasets used for $\mathcal{LM}$ training. These datasets contain instruction-response pairs derived from eight publicly available CXR datasets including MIMIC-CXR-JPG \cite{johnson2019mimic}, VinDr-CXR \cite{nguyen2022vindr}, MS-CXR \cite{boecking2022making}, PadChest-Grounding \cite{de2025padchest}, SLAKE \cite{liu2021slake}, MIMIC-CXR-VQA \cite{bae2024mimic}, RaDialog-Instruct \cite{pellegriniradialog} and Chest-ImaGenome \cite{wu2021chest}. VinDr-Instruct, and Anatomy Grounding instruction-response datasets were created using the VinDr-CXR \cite{nguyen2022vindr} and Chest ImaGenome \cite{wu2021chest} datasets, respectively. Details on these datasets creation are given in the supp. sec. S3. \\

\noindent \textbf{Radiology Tasks:} AnatomiX is trained on a diverse set of nine CXR-related tasks, spanning four categories: image understanding, grounding, report generation, and visual question answering (VQA), as illustrated in Fig.~\ref{fig:fig_dataset_tasks}. Details are provided in the supp. material section S4. \\

\noindent \textbf{Training Scheme:} AnatomiX is trained in three steps that focus on different architectural components and tasks. The first step focuses on the end-to-end training of APM for anatomical object detection and contrastive alignment. We set the number of anatomical objects ($N$) = 36 and train APM for 30 epochs with $1\times e^{-4}$  learning rate. Followed by an alignment step (step 2), where we align the embedding space of $\mathcal{LM}$ and APM by unfreezing the $\mathcal{P}_{im}$ and $\mathcal{P}_{A}$ projectors while keeping all other components frozen. This step uses the report generation dataset for 2 epoch training with $2\times e^{-4}$ learning. The third and final step focuses on instruction tuning, where we train $\mathcal{LM}$ (using LoRA \cite{hu2022lora}) along with $\mathcal{P}_{im}$ and $\mathcal{P}_{A}$ for 3 epochs while keeping APM frozen. This step includes supervised fine-tuning on all nine tasks discussed in supp. material section S4. All training steps use AdamW \cite{loshchilov2017decoupled} optimizer and were trained on 4 NVIDIA H100 GPUs with 80GB memory.

%% file: sec/results.tex
\begin{figure*}[ht]
  \centering
  \includegraphics[width=0.9\linewidth]{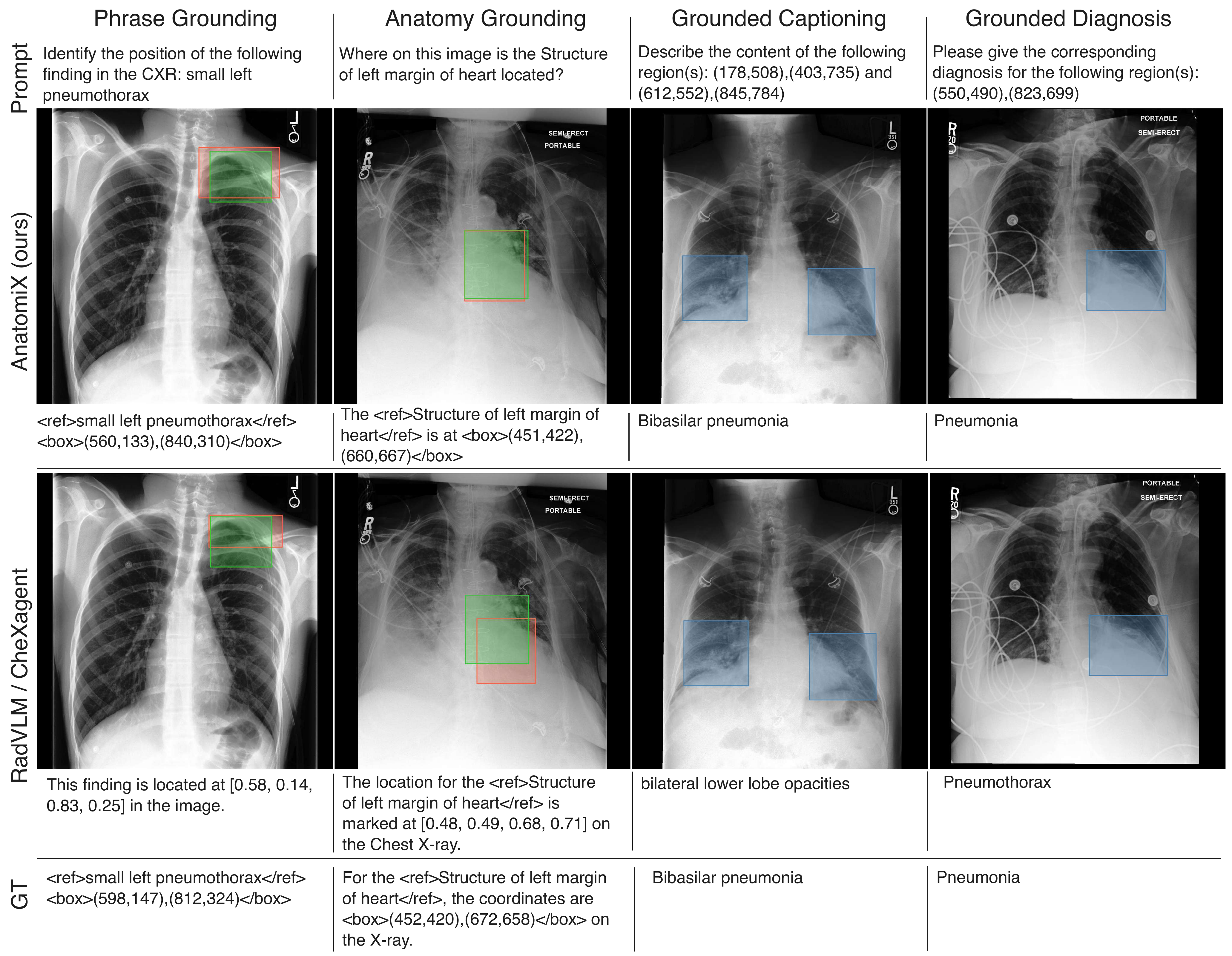}
  \caption{Sample input-output pairs and comparison with second best models on grounding tasks. The upper panels show outputs from our model across four tasks. The lower panels compare AnatomiX with RadVLM \cite{deperrois2025radvlm} for phrase and anatomy grounding, and with CheXagent \cite{chen2024chexagent} for grounded diagnosis and captioning. Box colors: blue = user input, green = ground truth, red = model output.}

  \label{fig:grounding_tasks}
\end{figure*}


\section{Results and Discussion}

\noindent \textbf{Grounding:} We begin by evaluating the model’s visual grounding capabilities. Specifically, we examine its ability to highlight relevant regions or pathologies on CXRs and to describe the features present within those regions. This information allows clinicians to visually verify the model’s predictions and gain insight into its decision making process. To assess these capabilities, we evaluate AnatomiX on four challenging grounding tasks using a combination of natural language generation (NLG), clinical, and detection metrics. We primarily compare our model against existing MLLMs with grounding capabilities, including RadVLM \cite{deperrois2025radvlm}, Maira-2 \cite{bannur2024maira}, and CheXagent \cite{chen2024chexagent}. For each compared model, we strictly follow the recommended input-output box coordinates processing scheme for fair comparison.

The phrase grounding and anatomy grounding tasks require localizing entities using bounding boxes. Accordingly, we evaluate these tasks using IoU and mean average precision (mAP) metrics. As shown in Table \ref{tab:grounding-tasks}, AnatomiX significantly outperforms all other models by up to 15\% in phrase grounding and over 25\% in anatomy grounding. This substantial improvement stems from the anatomy-oriented design of AnatomiX, which enables the model to focus more effectively on specific anatomical structures. Fig. \ref{fig:grounding_tasks} illustrates representative samples of phrase and anatomy grounding tasks, where our model accurately localizes both anatomical regions and pathologies compared to the second best models RadVLM \cite{deperrois2025radvlm} and CheXagent \cite{chen2024chexagent}. Detailed anatomy-wise results for anatomy grounding task are given in the supp. Tables S9-S12 (a).

\begin{figure*}[ht]
  \centering
  \includegraphics[width=0.9\linewidth]{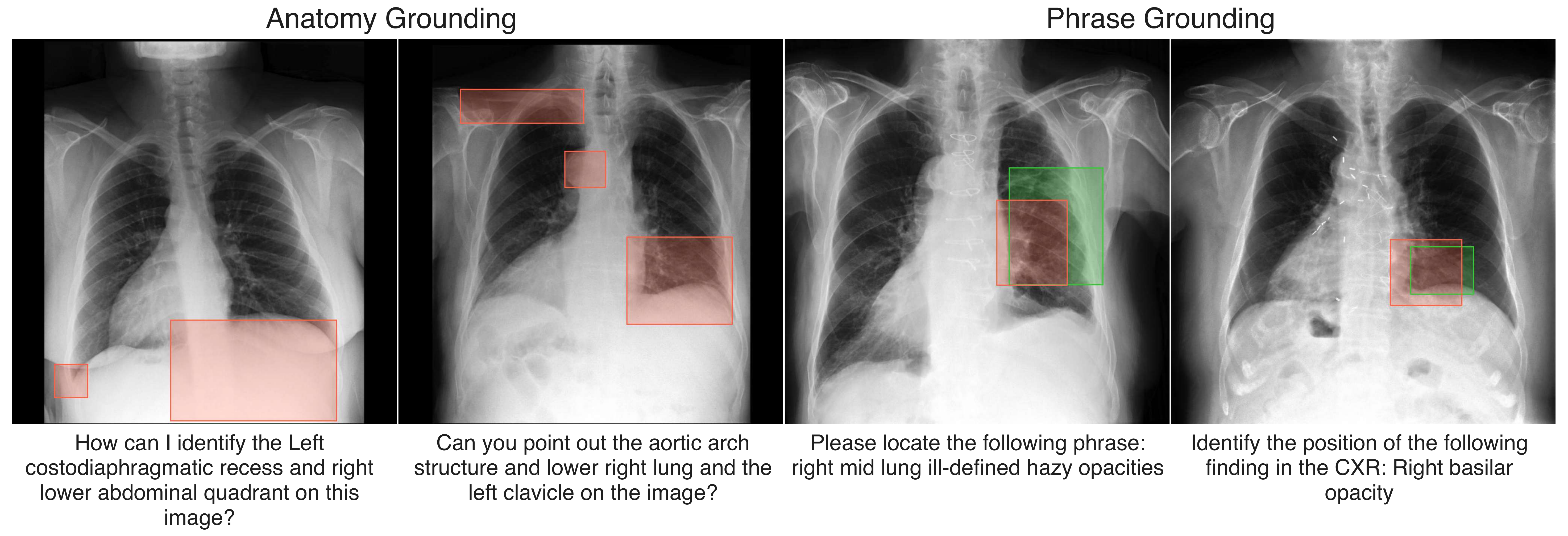}
    \caption{AnatomiX output for anatomy and phrase grounding on flipped images (left $\leftrightarrow$ right) with radiographic markers removed.}
   \label{fig:anatomy_grounding_wo_markers_1}
\end{figure*}

To further evaluate anatomical understanding, we conduct phrase and anatomy grounding tasks on horizontally flipped images, as illustrated in Fig. \ref{fig:fig_left_right} (c–e). RadVLM \cite{deperrois2025radvlm}, the best-performing baseline model, achieves strong results on standard inputs but fails on flipped images, frequently confusing left–right structures and relying on orientation cues. In contrast, AnatomiX maintains accurate grounding under image flipping, demonstrating robust and spatially consistent anatomical reasoning, which is essential for reliable medical image interpretation. Supp. Tables S9-S12 (b–c) present detailed quantitative results, showing that RadVLM performs drops drastically on anatomical objects requiring correct laterality identification e.g., left/right/upper/lower (Tables S10 and S11), with an average overall IoU/mAP of 0.108/0.08. On the contrary, AnatomiX shows no performance degradation in this challenging setting and achieves an average IoU/mAP of 0.712/0.605. Since DL models are prone to shortcut learning and may rely on radiological markers in chest X-rays to infer orientation, we conducted an additional experiment in which radiological markers (e.g., text labels or AP/PA indicators) were manually removed from a subset of images. Manual removal was performed to prevent visual artifacts that could arise from automated editing methods and unintentionally influence model predictions; consequently, only a subset of samples was used in this experiment. As shown in Fig. \ref{fig:anatomy_grounding_wo_markers_1} and supp. Fig. S1, AnatomiX continues to correctly localize diverse anatomical structures and phrases even in the absence of these markers, confirming that its performance is not driven by superficial orientation cues but by genuine anatomical understanding.

The grounded diagnosis and grounded captioning tasks, in contrast, require the model to identify pathologies or describe image content within a user-specified region. We evaluate performance on these tasks using both NLG (ROUGE \cite{lin2004rouge}, BERTScore \cite{zhang2019bertscore}, and METEOR \cite{banerjee2005meteor}) and clinical (RadGraph-F1 \cite{delbrouck2022improving} and CheXbert-14-F1 \cite{smit2020chexbert}) metrics. As shown in Table \ref{tab:grounding-tasks}, AnatomiX consistently achieves the highest scores across all metrics providing upto 30\% gains in grounded diagnosis and over 25 \% improvement in grounded captioning tasks, further underscoring the effectiveness of its anatomy-aware architecture. Fig. \ref{fig:grounding_tasks} show sample input-outputs of AnatomiX and the second best model CheXagent \cite{chen2024chexagent} for these two tasks, showcasing our model's capacity to generate precise and clinically meaningful descriptions aligned with the specified regions. Notably, MAIRA-2 \cite{bannur2024maira} model completely fails to perform these grounding tasks, as it was not trained to incorporate spatial or region-specific input. \\

\setlength{\tabcolsep}{3pt} 
\begin{table}[b]
\caption{NLG and Clinical metrics for Report Generation task.
}
\centering
\scriptsize
\adjustbox{width=\columnwidth}{
\begin{tabular}{l|ccc|cc}
\toprule
\textbf{Model} &
\multicolumn{3}{c|}{\textbf{NLG Metrics}} &
\multicolumn{2}{c}{\textbf{Clinical Metrics}} \\
\cmidrule(lr){2-4} \cmidrule(lr){5-6}
& ROUGE &BERTScore & METEOR &
   RadGraph & CheXbert-14 F1 \\
\midrule
MAIRA-2  & 0.43 & 0.25 & 0.12 & 0.17 & \underline{0.45} \\ 
Radialog & \underline{0.51} & \underline{0.35} & \underline{0.18} & \underline{0.24} & \textbf{0.48} \\
MedGamma & 0.37 & 0.29 & 0.18 & 0.20 & 0.40 \\
RadVLM   & 0.45 & 0.27 & 0.12 & 0.19 & 0.32 \\
CheXagent& 0.32 & 0.16 & 0.06 & 0.15 & 0.31 \\
\textbf{AnatomiX (ours)} & \textbf{0.53} & \textbf{0.38} & \textbf{0.21} & \textbf{0.26} & 0.42 \\

\bottomrule
\end{tabular}
}
\label{tab:result_report_gen}
\end{table}



\begin{table}[t]
\caption{Performance on image classification and detection tasks.}
\centering
\scriptsize
\setlength{\tabcolsep}{6.4pt}
\adjustbox{width=\columnwidth}{
\begin{tabular}{
l|
*{1}{>{\centering\arraybackslash}p{0.1\textwidth}}|
*{2}{>{\centering\arraybackslash}p{0.08\textwidth}}
}
\toprule
\textbf{Model} &
\multicolumn{1}{c|}{\textbf{Classification}} &
\multicolumn{2}{c}{\textbf{Detection}} \\
\cmidrule(lr){2-2}\cmidrule(lr){3-4} & CheXbert-14 F1
& IoU & mAP \\
\midrule
MAIRA-2   & 0.00 & 0.16 & 0.01 \\
Radialog  & 0.47 & 0.00 & 0.00 \\
MedGemma  & 0.40 & 0.04 & 0.00 \\
RadVLM    & 0.43 & 0.28 & 0.12 \\
CheXagent & \underline{0.85} & \underline{0.31} & \textbf{0.22} \\
\textbf{AnatomiX (ours)} & \textbf{0.85} & \textbf{0.31} & \underline{0.20} \\

\bottomrule
\end{tabular}
}
\label{tab:results_image_understanding}
\end{table}


\noindent \textbf{Report Generation:} The automatic generation of radiology reports is an important task that significantly reduces the time required for CXR interpretation and reporting. We evaluate this task using a set of NLG (ROUGE \cite{lin2004rouge}, BERTScore \cite{zhang2019bertscore}, METEOR \cite{banerjee2005meteor}) and clinical (RadGraph-F1 \cite{delbrouck2022improving}, and CheXbert-14-F1 \cite{smit2020chexbert}) metrics. We compare our model against several SOTA CXR report generation models, including RadVLM \cite{deperrois2025radvlm}, Maira-2 \cite{bannur2024maira}, CheXagent \cite{chen2024chexagent}, Radialog \cite{pellegriniradialog}, and Medgemma \cite{sellergren2025medgemma}. Table \ref{tab:result_report_gen} shows that AnatomiX consistently outperforms competing approaches across metrics, demonstrating its strong capability in producing both linguistically coherent and clinically accurate reports. The only exception occurs in CheXbert-14-F1 \cite{smit2020chexbert}, where AnatomiX attains an F1 score of 0.42, compared to 0.48 for Radialog and 0.45 for Maira-2. Importantly, both of these models contain approximately 1.5 $\times$ more parameters than AnatomiX, emphasizing the efficiency and scalability of our approach. These results collectively highlight AnatomiX’s balance between performance and computational efficiency, suggesting that it effectively captures domain-specific medical semantics without relying on excessively large model architectures. \\




\noindent \textbf{Image Understanding and Visual Question Answering (VQA):} For image understanding tasks (image classification and abnormality detection) we evaluate using CheXbert-14-F1 \cite{smit2020chexbert}, IoU, and mAP. AnatomiX outperforms all compared models in classification, achieves the highest IoU for abnormality detection, and maintains competitive mAP performance, highlighting strong visual reasoning and reliable localization (Table~\ref{tab:results_image_understanding}). For VQA, we benchmark on open- and close-ended tasks using BERTScore \cite{zhang2019bertscore} and CheXbert-14-F1. AnatomiX demonstrates strong performance across both settings (Table \ref{tab:vqa-results}), outperforming all models and matching or exceeding CheXagent \cite{chen2024chexagent}. Minor score differences mainly arise from keyword mismatches with ground truth—for example, when the answer to \textit{“Is there any pneumonia in the image?”} is \textit{“yes, pneumonia is present”} but the model outputs \textit{“yes”}, or vice versa—leading to lower metric scores despite clinical correctness. Overall, these results highlight AnatomiX’s strong image understanding and high VQA capabilities.



\begin{table}[t]
\caption{Open and close ended VQA task performance.}
\centering
\scriptsize
\setlength{\tabcolsep}{3pt}
\adjustbox{width=\columnwidth}{
\begin{tabular}{l|cc|cc}
\toprule
\multirow{2}{*}{\textbf{Model}} &
\multicolumn{2}{c|}{\textbf{Open-Ended VQA}} &
\multicolumn{2}{c}{\textbf{Close-Ended VQA}} \\
\cmidrule(lr){2-3} \cmidrule(lr){4-5}
 & \multicolumn{1}{c}{BERTScore} &
   \multicolumn{1}{c}{CheXbert-14 F1} &
   \multicolumn{1}{c}{BERTScore} &
   \multicolumn{1}{c}{CheXbert-14 F1} \\
\midrule
MAIRA-2    & 0.07 & 0.31 & 0.10 & 0.81 \\
Radialog   & 0.08 & 0.43 & 0.03 & 0.92 \\
MedGemma   & 0.03  & 0.44 & 0.02 & 0.38 \\
RadVLM     & 0.07 & 0.04 & 0.23 & 0.67 \\
CheXagent  & \underline{0.86} & \textbf{0.87} & \textbf{0.90} & \textbf{0.97} \\
\textbf{AnatomiX (ours)} & \textbf{0.86} & \underline{0.86} & \underline{0.89} & \underline{0.95} \\

\bottomrule
\end{tabular}
}
\label{tab:vqa-results}
\end{table}

%% file: sec/conclusion.tex
\section{Conclusion and Future Work}
In conclusion, AnatomiX shows significant improvements in CXR interpretation, especially in tasks that require a direct anatomical understanding. Our results highlight that anatomy-oriented design is the key to accurate spatial reasoning in medical MLLMs. While simple finetuning natural image MLLMs on large medical datasets can help, it can create false spatial correspondences. Future work could extend anatomy oriented architectures to other modalities such as MRI and improve the current architecture by reducing the potential redundancy multi-modal prompt. Furthermore, this study focuses on single turn interactions; extending it to multi turn setups would enhance flexibility and applicability. Overall, this work marks an important step toward domain-specific design in MLLMs.\\

\noindent \textbf{Acknowledgment:} This work was supported in part through the Minerva computational and data resources \cite{kovatch2020optimizing} and staff expertise provided by Scientific Computing and Data at the Icahn School of Medicine at Mount Sinai and supported by the Clinical and Translational Science Awards (CTSA) grant UL1TR004419 from the National Centre for Advancing Translational Sciences.

%% file: sec/X_suppl.tex



\clearpage
\onecolumn
\setcounter{page}{1}
\setcounter{section}{0}
\setcounter{figure}{0}
\setcounter{table}{0}
\setcounter{equation}{0}
\renewcommand\thesection{S\arabic{section}}
\renewcommand\thefigure{S\arabic{figure}}
\renewcommand\thetable{S\arabic{table}}
\renewcommand\theequation{S\arabic{equation}}

\begin{center}
    \textbf{\Large AnatomiX, an Anatomy-Aware Grounded Multimodal Large Language Model}\\[0.5em]
    \textbf{\Large for Chest X-Ray Interpretation}\\[1.5em] 
    \Large Supplementary Material
\end{center}


\begin{figure}[h]
    \centering
    \includegraphics[width=0.98\linewidth]{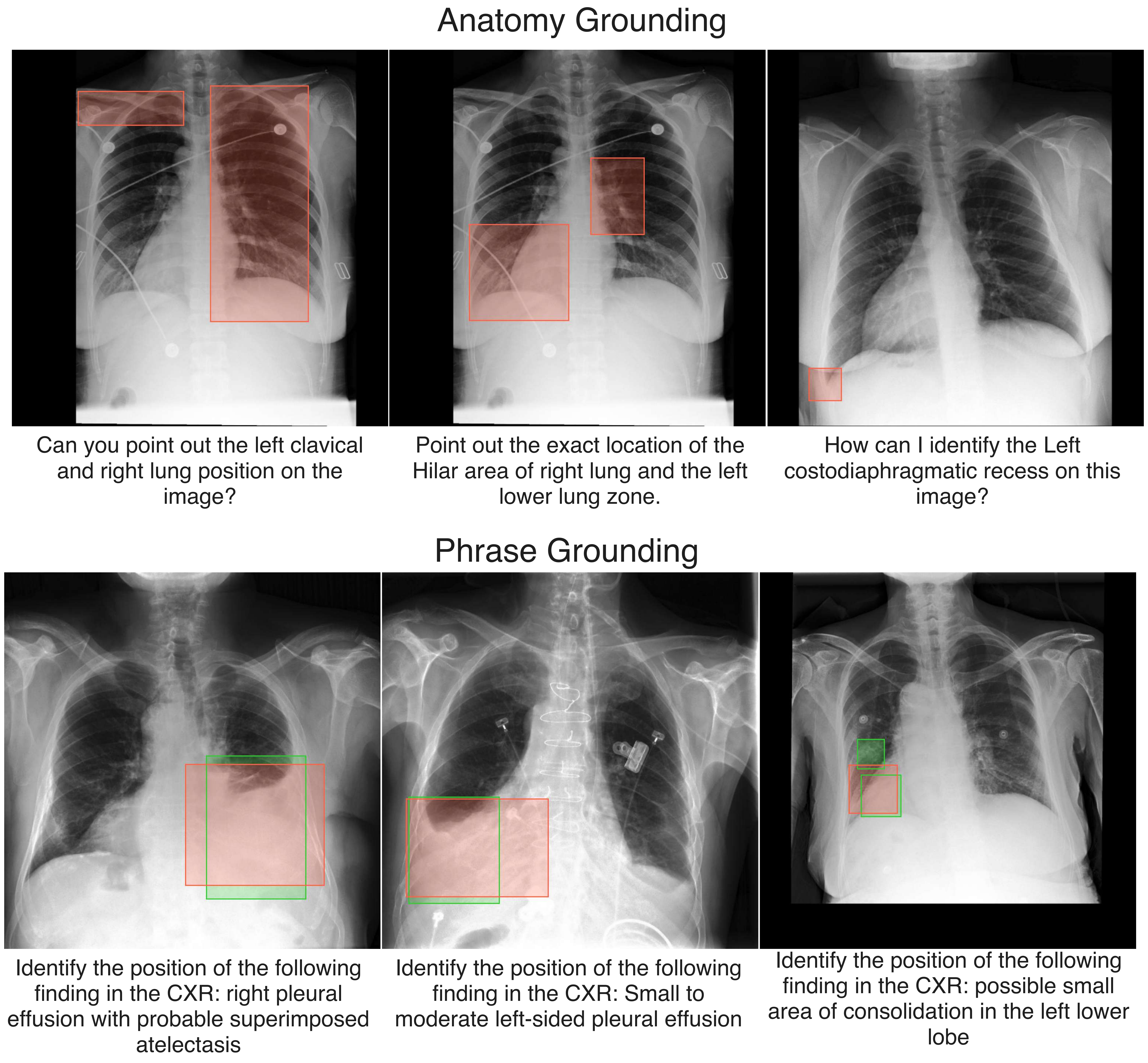}
    \caption{Additional samples for AnatomiX's output for anatomy and phrase grounding on flipped images (left $\leftrightarrow$ right) with radiographic markers removed. Bottom: Green boxes show the ground truth, while red show the model prediction.}    \label{fig:supp_phrase_and_grounding_wo_markers}
\end{figure}

\begin{figure}[H]
\centering
\begin{tcolorbox}[
    width=\linewidth,
    colframe=teal!90!green,      
    colback=white!100!green,      
    coltitle=white,             
    colbacktitle=teal!90!green,  
    title={Multimodal Prompt},
    boxrule=0.8pt,
]

\small

\textbf{User:}\\
You are a professional radiologist. I will provide you with context containing likely features about different parts of the chest X-rays.\\[0.5em]
\textbf{Image:}\\
\texttt{<image\_start> <image> … <image> <image\_end>}\\[0.2em]
\textbf{Likely findings:}\\
\texttt{\embtag{<emb>}<obj\_0>\embtag{</emb>} \boxtag{<box>}(0,387),(670,1024)\boxtag{</box>} Abdominal cavity shows enteric tube.}\\
\texttt{\embtag{<emb>}<obj\_1>\embtag{</emb>} \boxtag{<box>}(300,118),(394,207)\boxtag{</box>} Aortic arch structure is healthy.}\\
\texttt{...}\\
\texttt{\embtag{<emb>}<obj\_$N$>\embtag{</emb>} \boxtag{<box>}(398,391),(517,518)\boxtag{</box>} Right cardiophrenic sulcus is healthy.}\\[0.5em]
\textbf{Task:} \texttt{\brk{[TASK]}}\\[0.5em]
\textbf{Model Response:} \texttt{\brk{[MODEL RESPONSE]}}
\end{tcolorbox}

\caption{Multimodal prompt template used in $\mathcal{LM}$. Colored tags (\texttt{<emb>} and \texttt{<box>}) denote special tokens corresponding to anatomical object embeddings and bounding boxes, respectively. Each \texttt{<obj\_i>} token represents the embedding of the $i^{\text{th}}$ anatomical object, while \texttt{<image>} indicates image patch embeddings.}
\label{fig:fig_multimodal_prompt}
\end{figure}


\section{Self-Similarity Loss Matrix} \label{sec:supp_self_sim_loss}

The Contrastive Alignment stage of the Anatomy Perception Module (APM) utilizes the Self-Similarity matrix $\mathrm{S}_{self}$ to model fine grained semantic relations among anatomical descriptions. In this stage, a pretrained sentence encoder $\mathcal{S}$ provides indirect supervision by embedding textual inputs into a continuous semantic space that captures linguistic and clinical similarities. Given a set of input sentences $S_t$, each sentence is encoded through $\mathcal{S}$ to obtain text embeddings $\mathrm{S}_E \in \mathbb{R}^{N \times 768}$, where each row corresponds to the representation of one sentence in a 768-dimensional embedding space.
To ensure consistency and comparability across representations, the embeddings are first normalized using $\ell_2$ normalization:

\begin{equation}
    \bar{\mathrm{S}}_{E} = \frac{\mathrm{S}_E}{|\mathrm{S}_E|_2}
\end{equation}

This normalization projects the embeddings onto a unit hypersphere, ensuring that they encode directional (semantic) differences rather than magnitude based variations. The normalized embeddings are then used to compute the Self-Similarity matrix:

\begin{equation}
\mathrm{S}_{self} = \text{Softmax}(\bar{\mathrm{S}}_{E}) \cdot \text{Softmax}(\bar{\mathrm{S}}_{E})^{\mathrm{T}}
\end{equation}

\noindent where $\mathrm{S}_{self} \in \mathbb{R}^{N \times N}$ encodes pairwise similarity scores between all sentences in $S_t$. The softmax operation (applied row wise) ensures these similarities are smooth and probabilistically interpretable. 

Next, to align anatomical and textual semantics, we compute a projected similarity matrix $\hat{\mathrm{K}}_{A}$ between the projected anatomical features $\hat{\mathrm{O}}_{A}$ and projected text embeddings $\hat{\mathrm{S}}_A$:

\begin{equation}
\hat{\mathrm{K}}_{A} = \text{Softmax}\left( \frac{ \hat{\mathrm{O}}_{A} \hat{\mathrm{S}}_A^{\mathrm{T}} }{ \tau } \right)
\end{equation}

\noindent where $\hat{\mathrm{K}}_{A} \in \mathbb{R}^{N \times N}$, and $\tau$ is the temperature coefficient (set to $0.01$) that controls the sharpness of the similarity distribution.


The final contrastive alignment loss is defined as the averaged KL-divergence between the anatomical–textual similarity matrix $\hat{\mathrm{K}}_{A}$ and the self-similarity matrix $\mathrm{S}_{self}$, computed in both row-wise and column-wise directions to enforce mutual consistency:

\begin{align}
\mathcal{L}_{CL}
&= \frac{1}{2} KL\left( \hat{\mathrm{K}}_{A}, \mathrm{S}_{self} \right)
+ \frac{1}{2} KL\left( \hat{\mathrm{K}}_{A}^{\mathrm{T}}, \mathrm{S}_{self}^{\mathrm{T}} \right)
\end{align}

This formulation encourages $\hat{\mathrm{O}}_{A}$ and $\hat{\mathrm{S}}_A$ to maintain pairwise relationships that reflect the semantic structure captured in $\mathrm{S}_{self}$. As a result, the APM preserves semantic coherence and clinical consistency across related sentences, capturing overlapping anatomical features rather than enforcing strict one-to-one alignments.

\begin{figure}[t!]
  \centering
  \hfill
  \includegraphics[width=\linewidth]{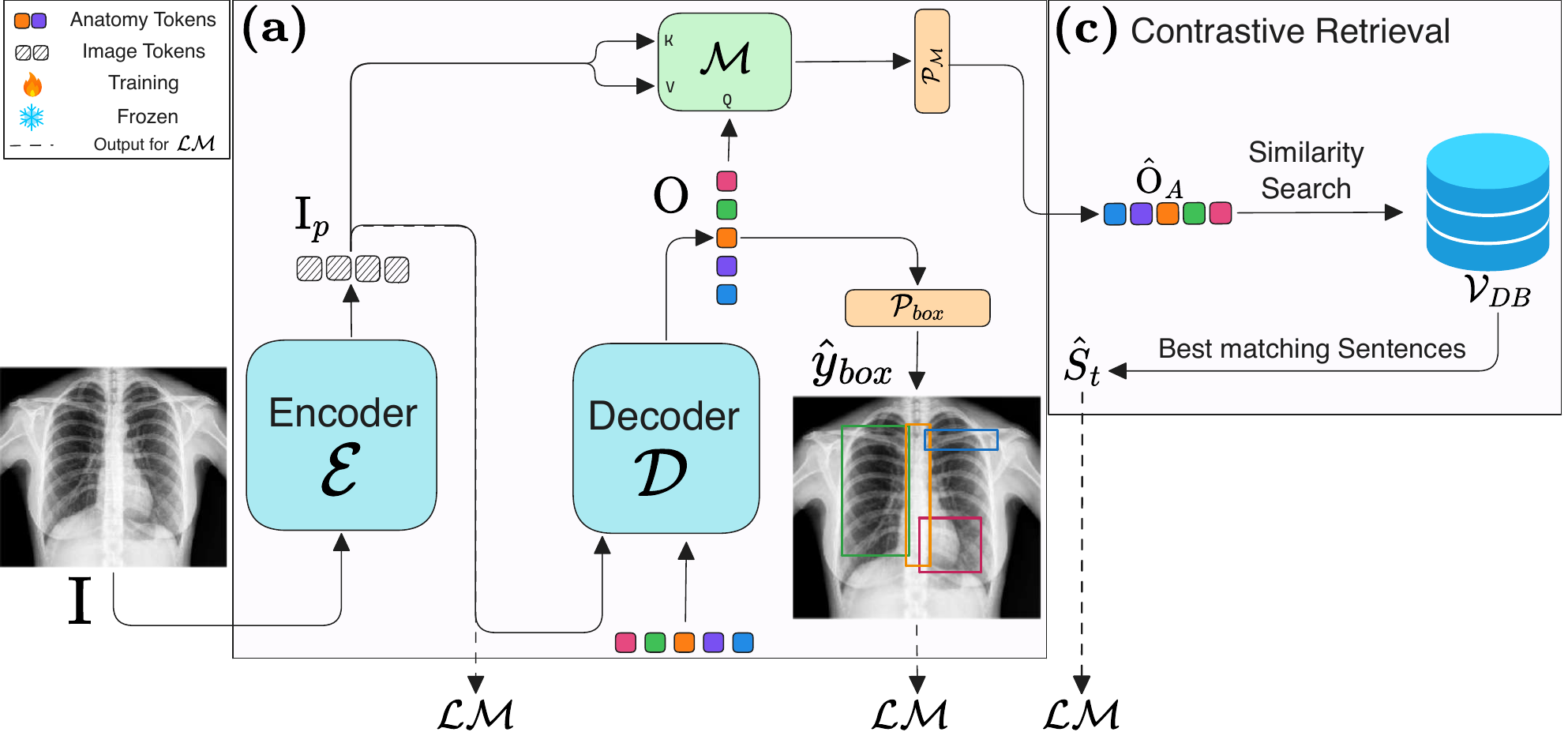}
  \caption{APM architecture during inference, where the Contrastive Alignment's components are replaced with vector database for the Contrastive Retrieval. See Fig. 2 for training architecture.}
  \label{fig:supp_fig_enc_dec_arch}
\end{figure}


\section{Vector Database} \label{sec:supp_databse}

During APM inference, we replace the Contrastive Alignment with the Contrastive Retrieval (see Fig. \ref{fig:supp_fig_enc_dec_arch}) to identify the semantically most similar sentences to the anatomical object tokens $\hat{\mathrm{O}}_A$. These retrieved sentences represent the most probable observations for each anatomical object and thus provide important contextual information for downstream descriptive tasks in $\mathcal{LM}$, as discussed in ablations.

\begin{figure}[H]
    \centering
    \includegraphics[width=\linewidth]{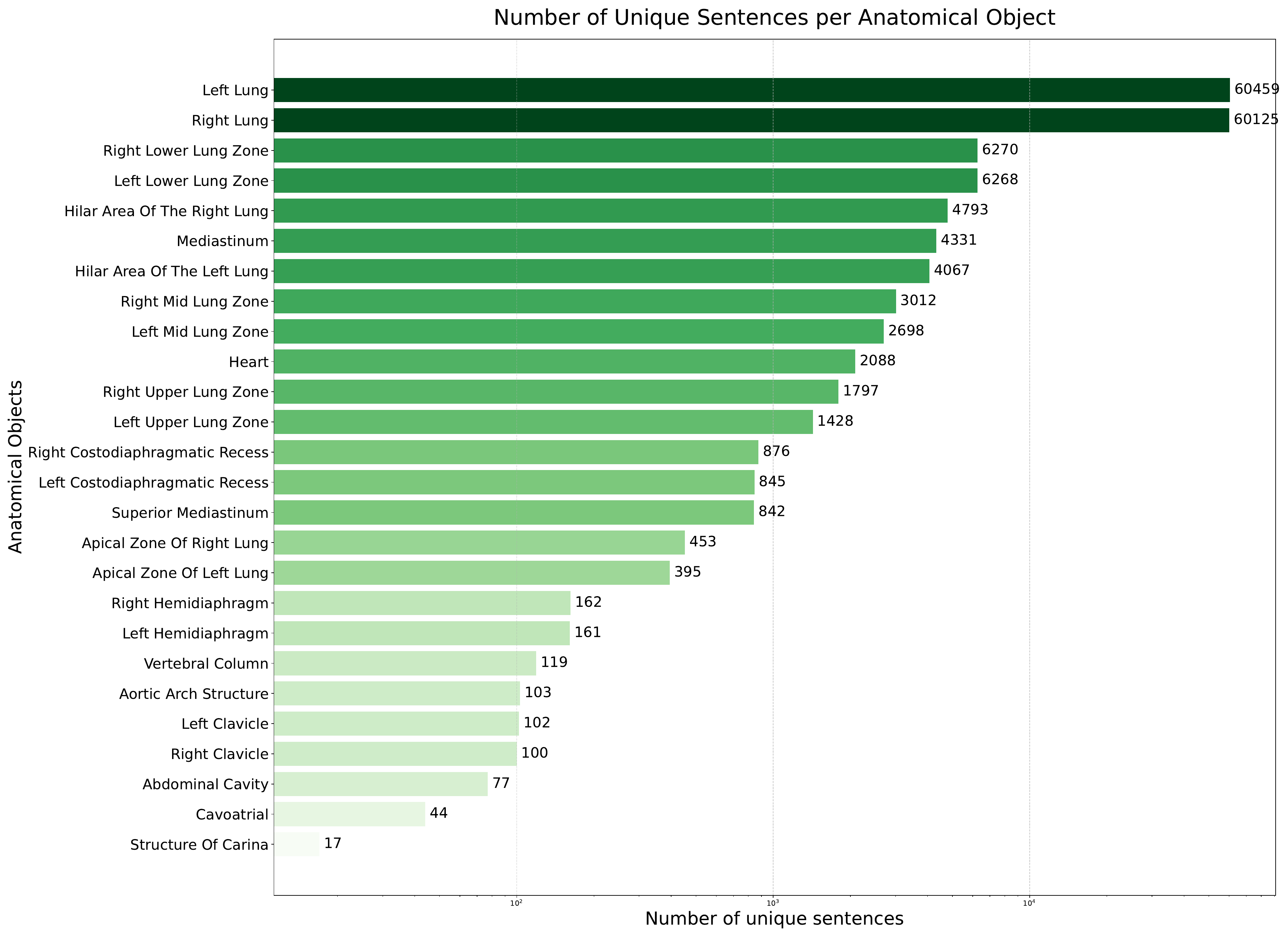}
    \caption{Size of the vector database. Each bar shows the number of unique sentences associated with a specific anatomical object. Anatomical objects with fewer than 10 unique sentences are omitted for clarity.}
    \label{fig:rag_db_dist}
\end{figure}

The vector database, denoted as $\mathcal{V}_{DB}$, stores all unique sentences associated with each anatomical object from the validation set of the Chest-ImaGenome dataset. For every object in an image, we construct a concise descriptive sentence using the corresponding phrases and attributes given in the original dataset (example sentence: “Right lower lung shows pleural effusion and atelectasis”). To build $\mathcal{V}_{DB}$, we first compile the set of unique sentences for each anatomical object. Each sentence is then encoded using the sentence encoder $\mathcal{S}$ and the trained projection head $\mathcal{P}_S$, producing $s$-dimensional embeddings (see Fig. 2 for $\mathcal{P}_S$). These embeddings, along with their corresponding sentences, are stored as key–value pairs in $\mathcal{V}_{DB}$, with a distinct sub-database allocated to each anatomical object. Consequently, $\mathcal{V}_{DB}$ comprises $N$ independent sub-databases. The full distribution for the size of each anaomical object database is shown in Fig. \ref{fig:rag_db_dist}.

The compact nature of both the embeddings and the sentences ensures that $\mathcal{V}_{DB}$ remains lightweight, enabling efficient retrieval at inference time. During inference, a similarity search is performed between each anatomical object token $\hat{\mathrm{O}}_A^i$ and the sentence embeddings within the corresponding sub-database of $\mathcal{V}_{DB}$, thereby retrieving the most relevant descriptive sentences for each object. These sentences are then passed to $\mathcal{LM}$ using a multimodal prompt template shown in Fig. \ref{fig:fig_multimodal_prompt}, where they provide important contextual information about each anatomy.

\section{Datasets} \label{sec:supp_datasets}

\subsection{Anatomy Grounding} 

For instruction tuning, we construct the Anatomy Grounding dataset using bounding box annotations for 36 distinct anatomical structures. We design 20 question and answer templates, as illustrated in Fig. \ref{fig:anatomy_grounding_templates}. The question templates query the location of a specific anatomical structure, while the answer templates include both the anatomical name and its location (i.e., bounding box coordinates). During dataset construction, we randomly sample from the question and answer templates to increase data diversity. The final dataset contains the same number of samples as Chest-ImaGenome, maintains a uniform distribution of anatomical structures, and follows the official data split of Chest-ImaGenome.

\subsection{VinDr-Instruct} 

The VinDr-Instruct dataset is constructed from the VinDr-CXR dataset and comprises question–answer pairs for abnormality detection, phrase grounding, and grounded diagnosis tasks. We adopt the question templates proposed in CheXagent, as illustrated in Fig. \ref{fig:vindr_template}. The answers are formatted as single- or multi-word responses that contain only the essential information, without full sentence structures. In this work, we follow the original train–test split of the VinDr-CXR dataset.

\begin{figure}[h]
\centering
\begin{tcolorbox}[
    width=\linewidth,
    colframe=teal!90!green,      
    colback=white!100!green,      
    coltitle=white,             
    colbacktitle=teal!90!green,  
    title={Templates for VinDr-Instruct},
    boxrule=0.8pt,
    boxsep=0.8pt,
]

{\fontsize{7.6}{8.5}\selectfont
\textbf{Questions:}
\begin{itemize} 
    \item Where is the \embtag{\{\texttt{anatomy}\}} located in this Chest X-ray?
    \item Can you point out the \embtag{\{\texttt{anatomy}\}}'s position on the image?
    \item What's the location of the \embtag{\{\texttt{anatomy}\}} in the X-ray?
    \item Identify where the \embtag{\{\texttt{anatomy}\}} is on this Chest X-ray, please.
    \item Where exactly is the \embtag{\{\texttt{anatomy}\}} found on this image?
    \item Could you specify where to find the \embtag{\{\texttt{anatomy}\}} on this X-ray?
    \item Highlight the \embtag{\{\texttt{anatomy}\}}'s area on the image.
    \item Show me the \embtag{\{\texttt{anatomy}\}}'s location on this CXR.
    \item Where should I look to find the \embtag{\{\texttt{anatomy}\}} in this image?
    \item Can you locate the \embtag{\{\texttt{anatomy}\}} on this X-ray for me?
    \item Please point to the \embtag{\{\texttt{anatomy}\}} on this Chest X-ray.
    \item Indicate the position of the \embtag{\{\texttt{anatomy}\}} on this image.
    \item Describe the location of the \embtag{\{\texttt{anatomy}\}} on the X-ray.
    \item Where on this image is the \embtag{\{\texttt{anatomy}\}} located?
    \item Point out the exact location of the \embtag{\{\texttt{anatomy}\}} in the Chest X-ray.
    \item How can I identify the \embtag{\{\texttt{anatomy}\}} on this image?
    \item Where is the \embtag{\{\texttt{anatomy}\}} situated in this CXR?
    \item Can you highlight the \embtag{\{\texttt{anatomy}\}} on this image?
    \item Indicate where the \embtag{\{\texttt{anatomy}\}} is found on this X-ray.
    \item Describe where to find the \embtag{\{\texttt{anatomy}\}} on this Chest X-ray.
\end{itemize}

\textbf{Answers:}
\begin{itemize}
    \item The <ref>\embtag{\{\texttt{anatomy}\}}</ref> is located at the coordinates <box>\boxtag{\{\texttt{boxes}\}}</box> on the image.
    \item You'll find the <ref>\embtag{\{\texttt{anatomy}\}}</ref> at  <box>\boxtag{\{\texttt{boxes}\}}</box>  in the X-ray.
    \item The <ref>\embtag{\{\texttt{anatomy}\}}</ref> can be seen at  <box>\boxtag{\{\texttt{boxes}\}}</box>  on the Chest X-ray.
    \item The location of the <ref>\embtag{\{\texttt{anatomy}\}}</ref> is at  <box>\boxtag{\{\texttt{boxes}\}}</box>  on the image.
    \item For the <ref>\embtag{\{\texttt{anatomy}\}}</ref>, the coordinates are  <box>\boxtag{\{\texttt{boxes}\}}</box>  on the X-ray.
    \item The <ref>\embtag{\{\texttt{anatomy}\}}</ref> is situated at  <box>\boxtag{\{\texttt{boxes}\}}</box>  in the image.
    \item On the Chest X-ray, the <ref>\embtag{\{\texttt{anatomy}\}}</ref> is located at  <box>\boxtag{\{\texttt{boxes}\}}</box>
    \item The <ref>\embtag{\{\texttt{anatomy}\}}</ref> appears at the coordinates  <box>\boxtag{\{\texttt{boxes}\}}</box>  on the image.
    \item In the X-ray, the <ref>\embtag{\{\texttt{anatomy}\}}</ref> is identifiable at  <box>\boxtag{\{\texttt{boxes}\}}</box>.
    \item The location for the <ref>\embtag{\{\texttt{anatomy}\}}</ref> is marked at  <box>\boxtag{\{\texttt{boxes}\}}</box>  on the Chest X-ray.
    \item The <ref>\embtag{\{\texttt{anatomy}\}}</ref> is positioned at  <box>\boxtag{\{\texttt{boxes}\}}</box>  on the image.
    \item The area occupied by the <ref>\embtag{\{\texttt{anatomy}\}}</ref> is at  <box>\boxtag{\{\texttt{boxes}\}}</box>  in the X-ray.
    \item On the image, you can find the <ref>\embtag{\{\texttt{anatomy}\}}</ref> at  <box>\boxtag{\{\texttt{boxes}\}}</box>.
    \item The <ref>\embtag{\{\texttt{anatomy}\}}</ref>'s  location is at  <box>\boxtag{\{\texttt{boxes}\}}</box>  on the Chest X-ray.
    \item In terms of coordinates, the <ref>\embtag{\{\texttt{anatomy}\}}</ref> is found at  <box>\boxtag{\{\texttt{boxes}\}}</box>  on the image.
    \item Regarding the <ref>\embtag{\{\texttt{anatomy}\}}</ref>, it is located at  <box>\boxtag{\{\texttt{boxes}\}}</box>  on the X-ray.
    \item The <ref>\embtag{\{\texttt{anatomy}\}}</ref> specifically is at  <box>\boxtag{\{\texttt{boxes}\}}</box>  on the Chest X-ray.
    \item Concerning the <ref>\embtag{\{\texttt{anatomy}\}}</ref>, you will find it at  <box>\boxtag{\{\texttt{boxes}\}}</box>  in the image.
    \item The <ref>\embtag{\{\texttt{anatomy}\}}</ref> is at  <box>\boxtag{\{\texttt{boxes}\}}</box>  on the X-ray.
    \item For identifying the <ref>\embtag{\{\texttt{anatomy}\}}</ref>, look at  <box>\boxtag{\{\texttt{boxes}\}}</box>  on the Chest X-ray.
\end{itemize}
}

\end{tcolorbox}

\caption{Instruction QA templates for Anatomy Grounding dataset. \embtag{\{\texttt{anatomy}\}} and \boxtag{\{\texttt{boxes}\}} represents the Anatomy name and location (coordinates), respectively.}

\label{fig:anatomy_grounding_templates}
\end{figure}

\begin{figure}[H]
\centering
\begin{tcolorbox}[
    width=\linewidth,
    colframe=teal!90!green,      
    colback=white!100!green,      
    coltitle=white,             
    colbacktitle=teal!90!green,  
    title={Templates for VinDr-Instruct},
    boxrule=0.8pt,
]
{\fontsize{7.6}{8.5}\selectfont
\textbf{Abnormality Detection:}
\begin{itemize}
    \item Detect \embtag{\{\texttt{disease}\}} in the given image.
    \item Locate areas in the chest X-ray where \embtag{\{\texttt{disease}\}} are present, using bounding box coordinates
    \item Perform abnormality detection (in the bounding box format) for the given image.
    \item Find the locations of \embtag{\{\texttt{disease}\}} in the bounding box format for the given image.
    \item Locate \embtag{\{\texttt{disease}\}} for the given image.
    \item Examine the chest X-ray and mark the regions affected by \embtag{\{\texttt{disease}\}} with bounding boxes.
    \item Detect the following in the image: \embtag{\{\texttt{disease}\}}.
    \item Examine the image for regions affected by \embtag{\{\texttt{disease}\}}, and indicate their positions with bounding boxes.
    \item Perform detection for \embtag{\{\texttt{disease}\}}.
\end{itemize}

\textbf{Phrase Grounding:}
\begin{itemize}
    \item Detect \embtag{\{\texttt{disease}\}} in the given image.
    \item Locate areas in the chest X-ray where \embtag{\{\texttt{disease}\}} is present, using bounding box coordinates.
    \item Localize \embtag{\{\texttt{disease}\}} in the bounding box format for the given image.
    \item Find the locations of \embtag{\{\texttt{disease}\}} in the bounding box format for the given image.
    \item Locate \embtag{\{\texttt{disease}\}} for the given image.
    \item Examine the chest X-ray and mark the regions affected by \embtag{\{\texttt{disease}\}} with bounding boxes.
    \item Detect the following in the image: \embtag{\{\texttt{disease}\}}.
    \item Examine the image for regions affected by \embtag{\{\texttt{disease}\}}, and indicate their positions with bounding boxes.
    \item Perform detection for \embtag{\{\texttt{disease}\}}.
\end{itemize}

\textbf{Grounded Diagnosis:}
\begin{itemize}
    \item Please give the corresponding diagnosis for the following region(s): \boxtag{\{\texttt{boxes}\}}
    \item Provide a diagnosis based on the content of the following region(s): \boxtag{\{\texttt{boxes}\}}
\end{itemize}

}

\end{tcolorbox}

\caption{Instruction templates used for generating VinDr-Instruct dataset. \embtag{\{\texttt{disease}\}} and \boxtag{\{\texttt{boxes}\}} represents the input abnormality name and box coordinates, respectively.}
\label{fig:vindr_template}
\end{figure}

\begin{table}[H]
\caption{Number of training, validation, and test samples for the nine datasets used in $\mathcal{LM}$ training and validation. The Source column indicates the original public dataset used directly or as the basis for dataset creation.}
\centering
\begin{tabular}{l l r r r}
\toprule
\textbf{Dataset} & \textbf{Source} & \textbf{Test} & \textbf{Train} & \textbf{Val} \\
\midrule
MIMIC-VQA & MIMIC-CXR-VQA & 5,497 & 101,963 & 4,926 \\
RaDialog Instruct & RaDialog Instruct & 799 & 6,274 & 822 \\
SLAKE & SLAKE & 298 & 1,175 & 285 \\
Anatomy Grounding & Chest-ImaGenome & 3,403 & 237,938 & 1,959 \\
MS-CXR & MS-CXR & 528 & 2,445 & 507 \\
VinDr-Instruct & VinDr-CXR & 6,166 & 38,122 & 4,099 \\
PadChest-Grounding & PadChest-Grounding & 1,121 & 3,920 & 558 \\
MIMIC-CXR Classification & MIMIC-CXR & 2,957 & 182,425 & 1,666 \\
MIMIC Report Gen & MIMIC-CXR & 1,722 & 135,049 & 1,078 \\
\midrule
\textbf{Total} & \textbf{--} & \textbf{22,491} & \textbf{709,311} & \textbf{15,900} \\
\bottomrule
\end{tabular}
\label{tab:dataset-stats-sources}
\end{table}

\section{Radiology Tasks} \label{sec:supp_tasks}
AnatomiX is trained and evaluated on nine CXR-related tasks, spanning four categories: image understanding, grounding, report generation, and visual question answering (VQA). Each of these tasks is focused on specific aspect of the CXR interpretation and uses different dataset(s). \\

\noindent \textbf{Image Understanding:}  This category includes multi-label image classification across 14 classes using the MIMIC-CXR dataset, as well as CXR abnormality detection leveraging the VinDr-Instruct dataset. Fig. \ref{fig:fig_vqa_and_image_understanding} shows sample input-output samples for classification and abnormality detection tasks along with the output of our model. \\

\noindent \textbf{Grounding:} We include four challenging grounding tasks in this work, namely: Phrase Grounding, Grounded Diagnosis, Grounded Captioning, and Anatomy Grounding. In Phrase Grounding, the model identifies the spatial location of a given phrase within an input image, utilizing the MS-CXR, PadChest-Gr, and VinDr-Instruct datasets. Grounded Diagnosis and Grounded Captioning require the model to infer a diagnosis and generate a textual description for a specified image region, respectively; we use VinDr-Instruct and MS-CXR for Grounded Diagnosis and MS-CXR for Grounded Captioning. Finally, Anatomy Grounding uses the Anatomy-Grounding dataset to localize anatomical structures based on user-provided textual prompts. Fig. 5 shows sample input and output pairs for these tasks, along with the output of our model. \\

\noindent \textbf{Report Generation:} This task involves generating the full report, including both the findings and impression sections using MIMIC-CXR dataset. A sample image-report is shown in Fig. \ref{fig:fig_report_gen} along with the output of AnatomiX. \\

\noindent \textbf{Visual Question Answering:} The VQA category consists of open-ended and closed-ended question answering tasks, derived from a combination of the MIMIC-VQA, SLAKE, and Radialog-Instruct datasets. 


\input{sec/ablation}

\section{Additional Results}

\begin{figure}[H]
  \centering
  \includegraphics[width=0.85\linewidth]{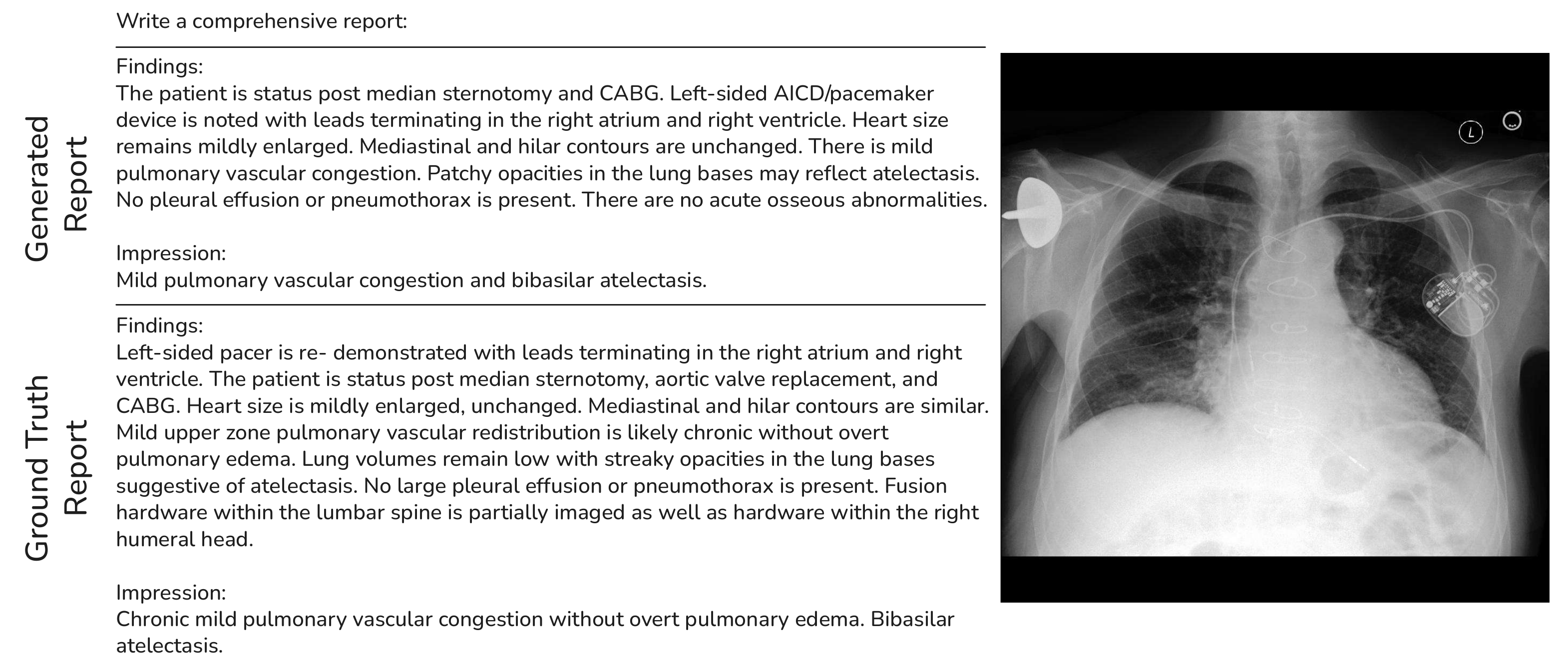}
  \caption{Sample report generation with AnatomiX.}
  \label{fig:fig_report_gen}
\end{figure}

\begin{figure}[H]
  \centering
  \includegraphics[width=0.85\linewidth]{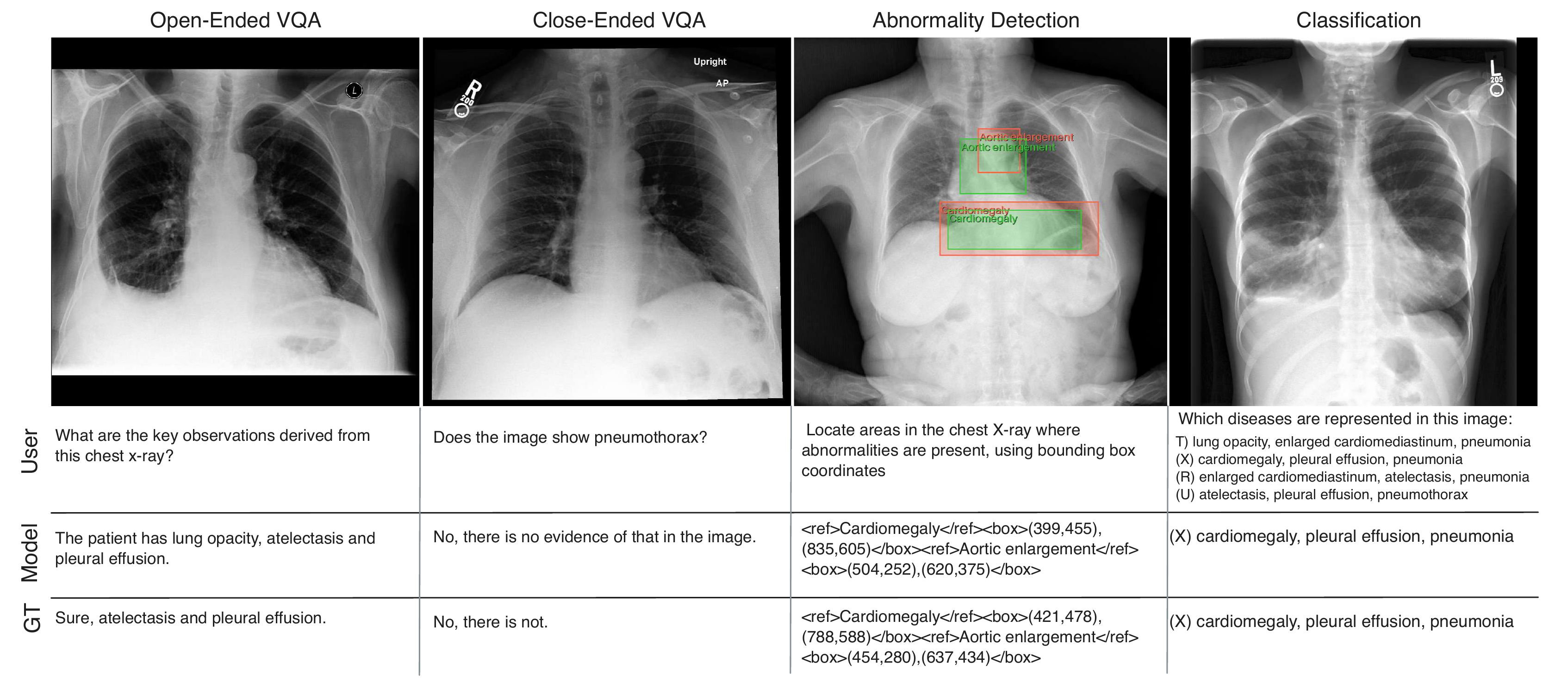}
  \caption{Example input-output-ground truth for image understanding and visual question answering tasks. Box colors: green represents the ground truth while red box shows the model's output.}
  \label{fig:fig_vqa_and_image_understanding}
\end{figure}

Table \ref{tab:phrase_grounding} contains the per dataset results for phrase grounding tasks, while Table \ref{tab:grounded_diagnosis} provides the per dataset breakdown for grounded diagnosis task. Furthermore, Fig. \ref{fig:fig_report_gen} shows a sample reporet generated using AnatomiX, and Fig. \ref{fig:fig_vqa_and_image_understanding} shows sample input-output for VQA and image understanding tasks.

\begin{table}[H]
\centering
\caption{Per dataset performance on Phrase Grounding task.}
\label{tab:phrase_grounding}
\begin{tabular}{lcc}
\toprule
\textbf{Dataset} & \textbf{IoU} & \textbf{mAP} \\
\midrule
MS-CXR      & 0.532 & 0.388 \\
PadChest-Gr & 0.444 & 0.337 \\
VinDr-Inst  & 0.230 & 0.180 \\
\bottomrule
\end{tabular}
\end{table}

\begin{table}[H]
\centering
\caption{Grounded diagnosis performance on different datasets.}
\label{tab:grounded_diagnosis}
\begin{tabular}{lccccc}
\toprule
\textbf{Dataset} & \textbf{BERTScore} & \textbf{ROUGE} & \textbf{METEOR} & \textbf{RadGraph-F1} & \textbf{CheXbert-F1} \\
\midrule
MS-CXR     & 0.758 & 0.729 & 0.402 & 0.732 & 0.744 \\
VinDr-Inst & 0.606 & 0.567 & 0.424 & 0.543 & 0.502 \\
\bottomrule
\end{tabular}
\end{table}


\begin{table*}[h]
\centering
\scriptsize
\setlength{\tabcolsep}{24pt}
\renewcommand{\arraystretch}{1.0}
\caption{Per-anatomy results for central anatomical structures with characteristic side predominance. (a) Results for the anatomy grounding task without flipping. (b–c) Anatomy grounding performance comparison between AnatomiX and RadVLM on flipped images. (d) Similarity between retrieved sentences $\hat{S}_t$ and ground-truth sentences in APM. }

\label{tab:mid_with_laterality_anatomy_structure_results}

\begin{adjustbox}{width=\textwidth}
\begin{tabular}{c|c c c c}

 & \rotatebox{90}{Heart}
 & \rotatebox{90}{Aortic arch structure}
 & \rotatebox{90}{Descending aorta}
 & \rotatebox{90}{Superior vena cava} \\ 
\hline

\multicolumn{5}{c}{\textbf{(a) AnatomiX: Anatomy Grounding in Normal Images (no flipping)}} \\
\hline
IoU  & 0.60 & 0.73 & 0.76 & 0.75 \\
mAP  & 0.71 & 0.63 & 0.64 & 0.65 \\

\hline
\multicolumn{5}{c}{\textbf{(b) AnatomiX: Anatomy Grounding in Horizontally Flipped Images}} \\
\hline
IoU  & 0.79 & 0.73 & 0.76 & 0.70 \\
mAP  & 0.69 & 0.61 & 0.64 & 0.57 \\

\hline
\multicolumn{5}{c}{\textbf{(c) RadVLM: Anatomy Grounding in Horizontally Flipped Images}} \\
\hline
IoU  & 0.59 & 0.05 & 0.04 & 0.06 \\
mAP  & 0.45 & 0.02 & 0.00 & 0.00 \\

\hline
\multicolumn{5}{c}{\textbf{(d) APM Sentence Retrieval Results}} \\
\hline
CheXbert-14 F1 & 0.92 & 0.41 & 1.00 & 1.00 \\
RadGraph F1    & 0.63 & 0.57 & 1.00 & 1.00 \\
METEOR         & 0.72 & 0.70 & 0.99 & 1.00 \\

\hline
\multicolumn{5}{c}{\textbf{(e) APM-wo-$\mathcal{M}$ Sentence Retrieval Results}} \\
\hline
CheXbert-14 F1 & 0.92 & 0.31 & 1.00 & 1.00 \\
RadGraph F1    & 0.60 & 0.51 & 1.00 & 1.00 \\
METEOR         & 0.71 & 0.66 & 0.99 & 1.00 \\

\hline
\multicolumn{5}{c}{\textbf{(f) APM-Dino Sentence Retrieval Results}} \\
\hline
CheXbert-14 F1 & 0.92 & 0.38 & 1.00 & 1.00 \\
RadGraph F1    & 0.61 & 0.57 & 1.00 & 1.00 \\
METEOR         & 0.70 & 0.68 & 0.99 & 1.00 \\

\hline
\multicolumn{5}{c}{\textbf{(g) APM-CLIP Sentence Retrieval Results}} \\
\hline
CheXbert-14 F1 & 0.92 & 0.25 & 1.00 & 1.00 \\
RadGraph F1    & 0.57 & 0.46 & 1.00 & 1.00 \\
METEOR         & 0.67 & 0.64 & 0.99 & 1.00 \\

\hline
\end{tabular}
\end{adjustbox}
\end{table*}


\begin{table}[!h]
\centering
\scriptsize
\setlength{\tabcolsep}{5pt}
\renewcommand{\arraystretch}{1.0}
\caption{Anatomy-wise results for left-sided anatomical structures. (a) Results for the anatomy grounding task without flipping. (b–c) Anatomy grounding performance comparison between AnatomiX and RadVLM on flipped images. (d) Similarity between retrieved sentences $\hat{S}_t$ and ground-truth sentences in APM. Note: Some anatomical objects achieve a perfect score of 1.0 due to the limited number of possible sentences.}
\label{tab:right_anatomy_structure_results}

\begin{adjustbox}{width=\textwidth}
\begin{tabular}{c|*{12}{c}}

 & \rotatebox{90}{Left lung}
 & \rotatebox{90}{Left upper lung zone}
 & \rotatebox{90}{Left mid lung zone}
 & \rotatebox{90}{Left lower lung zone}
 & \rotatebox{90}{Apical zone of left lung}
 & \rotatebox{90}{Hilar area of left lung}
 & \rotatebox{90}{Left costodiaphragmatic recess}
 & \rotatebox{90}{Left hemidiaphragm}
 & \rotatebox{90}{Left cardiophrenic sulcus}
 & \rotatebox{90}{Left clavicle}
 & \rotatebox{90}{Left upper abdominal quadrant}
 & \rotatebox{90}{Left margin of heart} \\

\hline
\multicolumn{13}{c}{\textbf{(a) AnatomiX: Anatomy Grounding in Normal Images (no flipping)}} \\
\hline
IoU 
& 0.88 & 0.86 & 0.77 & 0.80 & 0.81 & 0.67 & 0.61 & 0.74 & 0.60 & 0.74 & 0.79 & 0.62 \\

mAP
& 0.81 & 0.76 & 0.67 & 0.70 & 0.70 & 0.66 & 0.51 & 0.62 & 0.53 & 0.63 & 0.77 & 0.68 \\

\hline
\multicolumn{13}{c}{\textbf{(b) AnatomiX: Anatomy Grounding in Horizontally Flipped Images}} \\
\hline
IoU
& 0.85 & 0.84 & 0.73 & 0.79 & 0.81 & 0.78 & 0.60 & 0.72 & 0.60 & 0.54 & 0.84 & 0.80 \\

mAP
& 0.76 & 0.73 & 0.63 & 0.67 & 0.70 & 0.67 & 0.50 & 0.61 & 0.50 & 0.44 & 0.74 & 0.69 \\

\hline
\multicolumn{13}{c}{\textbf{(c) RadVLM: Anatomy Grounding in Horizontally Flipped Images}} \\
\hline
IoU 
& 0.01 & 0.00 & 0.00 & 0.00 & 0.00 & 0.00 & 0.00 & 0.00 & 0.00 & 0.00 & 0.01 & 0.04 \\

mAP 
& 0.00 & 0.00 & 0.00 & 0.00 & 0.00 & 0.00 & 0.00 & 0.00 & 0.00 & 0.00 & 0.00 & 0.00 \\

\hline
\multicolumn{13}{c}{\textbf{(d) APM Sentence Retrieval Results}} \\
\hline
CheXbert-14 F1 
& 0.58 & 0.11 & 0.15 & 0.31 & 0.32 & 0.39 & 0.34 & 0.85 & 1.00 & 0.75 & 1.00 & 1.00 \\

RadGraph F1 
& 0.49 & 0.49 & 0.45 & 0.55 & 0.70 & 0.57 & 0.71 & 0.76 & 1.00 & 0.64 & 1.00 & 1.00 \\

METEOR 
& 0.54 & 0.51 & 0.54 & 0.59 & 0.72 & 0.74 & 0.72 & 0.66 & 1.00 & 0.52 & 1.00 & 1.00 \\

\hline
\multicolumn{13}{c}{\textbf{(e) APM-wo-$\mathcal{M}$ Sentence Retrieval Results}} \\
\hline
CheXbert-14 F1 
& 0.57 & 0.06 & 0.09 & 0.29 & 0.59 & 0.37 & 0.45 & 0.71 & 1.00 & 0.61 & 1.00 & 1.00 \\

RadGraph F1 
& 0.48 & 0.46 & 0.42 & 0.54 & 0.84 & 0.58 & 0.77 & 0.70 & 1.00 & 0.55 & 1.00 & 1.00 \\

METEOR 
& 0.52 & 0.49 & 0.52 & 0.59 & 0.82 & 0.74 & 0.75 & 0.56 & 1.00 & 0.44 & 1.00 & 1.00 \\

\hline
\multicolumn{13}{c}{\textbf{(f) APM-Dino Sentence Retrieval Results}} \\
\hline
CheXbert-14 F1 
& 0.55 & 0.16 & 0.16 & 0.28 & 0.16 & 0.35 & 0.34 & 0.82 & 1.00 & 0.59 & 1.00 & 1.00 \\

RadGraph F1 
& 0.46 & 0.45 & 0.42 & 0.50 & 0.74 & 0.52 & 0.72 & 0.69 & 1.00 & 0.60 & 1.00 & 1.00 \\

METEOR 
& 0.51 & 0.50 & 0.52 & 0.57 & 0.70 & 0.73 & 0.71 & 0.54 & 1.00 & 0.44 & 1.00 & 1.00 \\

\hline
\multicolumn{13}{c}{\textbf{(g) APM-CLIP Sentence Retrieval Results}} \\
\hline
CheXbert-14 F1 
& 0.54 & 0.05 & 0.11 & 0.31 & 0.54 & 0.37 & 0.37 & 0.84 & 1.00 & 0.80 & 1.00 & 1.00 \\

RadGraph F1 
& 0.46 & 0.53 & 0.45 & 0.56 & 0.81 & 0.54 & 0.73 & 0.62 & 1.00 & 0.55 & 1.00 & 1.00 \\

METEOR 
& 0.50 & 0.50 & 0.51 & 0.58 & 0.80 & 0.73 & 0.72 & 0.46 & 1.00 & 0.46 & 1.00 & 1.00 \\

\hline
\end{tabular}
\end{adjustbox}
\end{table}


\begin{table*}[h]
\centering
\scriptsize
\setlength{\tabcolsep}{4pt}
\renewcommand{\arraystretch}{1.0}
\caption{Anatomy-wise results for right-sided anatomical structures. (a) Results for the anatomy grounding task without flipping. (b–c) Anatomy grounding performance comparison between AnatomiX and RadVLM on flipped images. (d) Similarity between retrieved sentences $\hat{S}_t$ and ground-truth sentences in APM. Note: Some anatomical objects achieve a perfect score of 1.0 due to the limited number of possible sentences.}
\label{tab:left_anatomy_structure_results}

\begin{adjustbox}{width=\textwidth}
\begin{tabular}{c|*{13}{c}}

 & \rotatebox{90}{Right lung}
 & \rotatebox{90}{Right upper lung zone}
 & \rotatebox{90}{Right mid lung zone}
 & \rotatebox{90}{Right lower lung zone}
 & \rotatebox{90}{Apical zone of right lung}
 & \rotatebox{90}{Hilar area of right lung}
 & \rotatebox{90}{Right costodiaphragmatic recess}
 & \rotatebox{90}{Right hemidiaphragm}
 & \rotatebox{90}{Right cardiophrenic sulcus}
 & \rotatebox{90}{Right clavicle}
 & \rotatebox{90}{Right upper abdominal quadrant}
 & \rotatebox{90}{Right atrial structure}
 & \rotatebox{90}{Right heart border} \\ \hline

\hline
\multicolumn{14}{c}{\textbf{(a) AnatomiX: Anatomy Grounding in Normal Images (no flipping)}} \\
\hline
IoU 
& 0.89 & 0.85 & 0.79 & 0.79 & 0.79 & 0.8 & 0.67 & 0.73 & 0.51 & 0.75 & 0.87 & 0.65 & 0.59 \\

mAP
& 0.82 & 0.76 & 0.68 & 0.7 & 0.68 & 0.71 & 0.57 & 0.63 & 0.42 & 0.63 & 0.77 & 0.57 & 0.64 \\

\hline
\multicolumn{14}{c}{\textbf{(b) AnatomiX: Anatomy Grounding in Horizontally Flipped Images}} \\
\hline
IoU
& 0.89 & 0.78 & 0.74 & 0.81 & 0.77 & 0.79 & 0.65 & 0.73 & 0.47 & 0.38 & 0.85 & 0.61 & 0.75 \\

mAP
& 0.81 & 0.68 & 0.62 & 0.7 & 0.67 & 0.68 & 0.55 & 0.61 & 0.35 & 0.27 & 0.74 & 0.51 & 0.63 \\

\hline
\multicolumn{14}{c}{\textbf{(c) RadVLM: Anatomy Grounding in Horizontally Flipped Images}} \\
\hline

IoU 
& 0.0 & 0.0 & 0.0 & 0.0 & 0.0 & 0.0 & 0.0 & 0.01 & 0.0 & 0.0 & 0.01 & 0.0 & 0.0 \\

mAP 
& 0.0 & 0.0 & 0.0 & 0.0 & 0.0 & 0.0 & 0.0 & 0.0 & 0.0 & 0.0 & 0.0 & 0.0 & 0.0 \\

\hline
\multicolumn{14}{c}{\textbf{(d) APM Sentence Retrieval Results}} \\
\hline

CheXbert-14 F1 
& 0.6 & 0.05 & 0.1 & 0.25 & 0.5 & 0.4 & 0.4 & 0.8 & 1.0 & 0.73 & 1.0 & 1.0 & 1.0 \\

RadGraph F1 
& 0.5 & 0.44 & 0.43 & 0.49 & 0.76 & 0.54 & 0.76 & 0.73 & 1.0 & 0.62 & 1.0 & 1.0 & 1.0 \\

METEOR 
& 0.55 & 0.5 & 0.53 & 0.57 & 0.75 & 0.74 & 0.74 & 0.61 & 1.0 & 0.49 & 1.0 & 1.0 & 1.0 \\

\hline
\multicolumn{14}{c}{\textbf{(e) APM-wo-$\mathcal{M}$ Sentence Retrieval Results}} \\
\hline

CheXbert-14 F1 
& 0.57 & 0.03 & 0.09 & 0.28 & 0.48 & 0.39 & 0.49 & 0.6 & 1.0 & 0.7 & 1.0 & 1.0 & 1.0 \\

RadGraph F1 
& 0.48 & 0.41 & 0.39 & 0.52 & 0.64 & 0.55 & 0.78 & 0.62 & 1.0 & 0.59 & 1.0 & 1.0 & 1.0 \\

METEOR 
& 0.53 & 0.48 & 0.52 & 0.58 & 0.74 & 0.74 & 0.76 & 0.47 & 1.0 & 0.43 & 1.0 & 1.0 & 1.0 \\

\hline
\multicolumn{14}{c}{\textbf{(f) APM-Dino Sentence Retrieval Results}} \\
\hline

CheXbert-14 F1 
& 0.57 & 0.05 & 0.1 & 0.24 & 0.35 & 0.38 & 0.41 & 0.63 & 1.0 & 0.78 & 1.0 & 1.0 & 1.0 \\

RadGraph F1 
& 0.47 & 0.47 & 0.37 & 0.47 & 0.74 & 0.51 & 0.73 & 0.61 & 1.0 & 0.61 & 1.0 & 1.0 & 1.0 \\

METEOR 
& 0.53 & 0.5 & 0.51 & 0.56 & 0.72 & 0.72 & 0.72 & 0.43 & 1.0 & 0.5 & 1.0 & 1.0 & 1.0 \\

\hline
\multicolumn{14}{c}{\textbf{(g) APM-CLIP Sentence Retrieval Results}} \\
\hline

CheXbert-14 F1 
& 0.56 & 0.05 & 0.09 & 0.24 & 0.7 & 0.4 & 0.53 & 0.85 & 1.0 & 0.91 & 1.0 & 1.0 & 1.0 \\ 

RadGraph F1 
& 0.47 & 0.48 & 0.4 & 0.51 & 0.81 & 0.53 & 0.76 & 0.69 & 1.0 & 0.56 & 1.0 & 1.0 & 1.0 \\

METEOR 
& 0.53 & 0.5 & 0.51 & 0.57 & 0.79 & 0.73 & 0.75 & 0.57 & 1.0 & 0.45 & 1.0 & 1.0 & 1.0 \\

\bottomrule
\end{tabular}
\end{adjustbox}

\end{table*}


\begin{table*}[h]
\centering
\scriptsize
\setlength{\tabcolsep}{11pt}
\renewcommand{\arraystretch}{1.0}

\caption{Anatomy-wise results for midline (central) anatomical structures. (a) Results for the anatomy grounding task without flipping. (b–c) Anatomy grounding performance comparison between AnatomiX and RadVLM on flipped images. (d) Similarity between retrieved sentences $\hat{S}_t$ and ground-truth sentences in APM. }
\label{tab:mid_anatomy_structure_results}
\begin{adjustbox}{width=\textwidth}
\begin{tabular}{c|ccccccc}

 & \rotatebox{90}{Trachea \& main bronchus}
 & \rotatebox{90}{Carina}
 & \rotatebox{90}{Mediastinum}
 & \rotatebox{90}{Superior mediastinum}
 & \rotatebox{90}{Vertebral column}
 & \rotatebox{90}{Cavoatrial}
 & \rotatebox{90}{Abdominal cavity} \\ \hline

\hline
\multicolumn{8}{c}{\textbf{(a) AnatomiX: Anatomy Grounding in Normal Images (no flipping)}} \\
\hline
IoU 
& 0.74 & 0.48 & 0.58 & 0.77 & 0.85 & 0.66 & 0.86 \\

mAP
& 0.62 & 0.40 & 0.73 & 0.67 & 0.74 & 0.57 & 0.81 \\

\hline
\multicolumn{8}{c}{\textbf{(b) AnatomiX: Anatomy Grounding in Horizontally Flipped Images}} \\
\hline
IoU
& 0.75 & 0.47 & 0.70 & 0.77 & 0.76 & 0.38 & 0.84 \\

mAP
& 0.64 & 0.35 & 0.59 & 0.66 & 0.65 & 0.29 & 0.77 \\

\hline
\multicolumn{8}{c}{\textbf{(c) RadVLM: Anatomy Grounding in Horizontally Flipped Images}} \\
\hline
IoU 
& 0.43 & 0.05 & 0.60 & 0.68 & 0.74 & 0.00 & 0.81 \\

mAP 
& 0.31 & 0.01 & 0.47 & 0.56 & 0.62 & 0.00 & 0.67 \\

\hline
\multicolumn{8}{c}{\textbf{(d) APM Sentence Retrieval Results}} \\
\hline
CheXbert-14 F1 
& 1.00 & 0.83 & 0.44 & 0.06 & 0.65 & 0.99 & 0.89 \\

RadGraph F1 
& 1.00 & 0.80 & 0.47 & 0.29 & 0.68 & 0.64 & 0.81 \\

METEOR 
& 0.99 & 0.86 & 0.45 & 0.44 & 0.62 & 0.57 & 0.82 \\

\hline
\multicolumn{8}{c}{\textbf{(e) APM-wo-$\mathcal{M}$ Sentence Retrieval Results}} \\
\hline
CheXbert-14 F1 
& 1.00 & 0.91 & 0.43 & 0.09 & 0.69 & 0.99 & 0.88 \\

RadGraph F1 
& 1.00 & 0.87 & 0.43 & 0.18 & 0.60 & 0.47 & 0.80 \\

METEOR 
& 0.99 & 0.91 & 0.43 & 0.45 & 0.61 & 0.32 & 0.83 \\

\hline
\multicolumn{8}{c}{\textbf{(f) APM-Dino Sentence Retrieval Results}} \\
\hline
CheXbert-14 F1 
& 1.00 & 0.89 & 0.40 & 0.05 & 0.57 & 0.99 & 0.88 \\

RadGraph F1 
& 1.00 & 0.88 & 0.44 & 0.21 & 0.55 & 0.50 & 0.82 \\

METEOR 
& 0.99 & 0.92 & 0.42 & 0.39 & 0.59 & 0.45 & 0.82 \\

\hline
\multicolumn{8}{c}{\textbf{(g) APM-CLIP Sentence Retrieval Results}} \\
\hline
CheXbert-14 F1 
& 1.00 & 0.79 & 0.44 & 0.02 & 0.59 & 0.99 & 0.78 \\ 

RadGraph F1 
& 1.00 & 0.75 & 0.37 & 0.22 & 0.64 & 0.44 & 0.58 \\

METEOR 
& 0.99 & 0.82 & 0.37 & 0.43 & 0.55 & 0.22 & 0.59 \\

\bottomrule
\end{tabular}
\end{adjustbox}

\end{table*}


\clearpage
\section{Attention Visualization} \label{sec:attention_maps}
\begin{figure}[h]
    \centering
    \includegraphics[width=\linewidth]{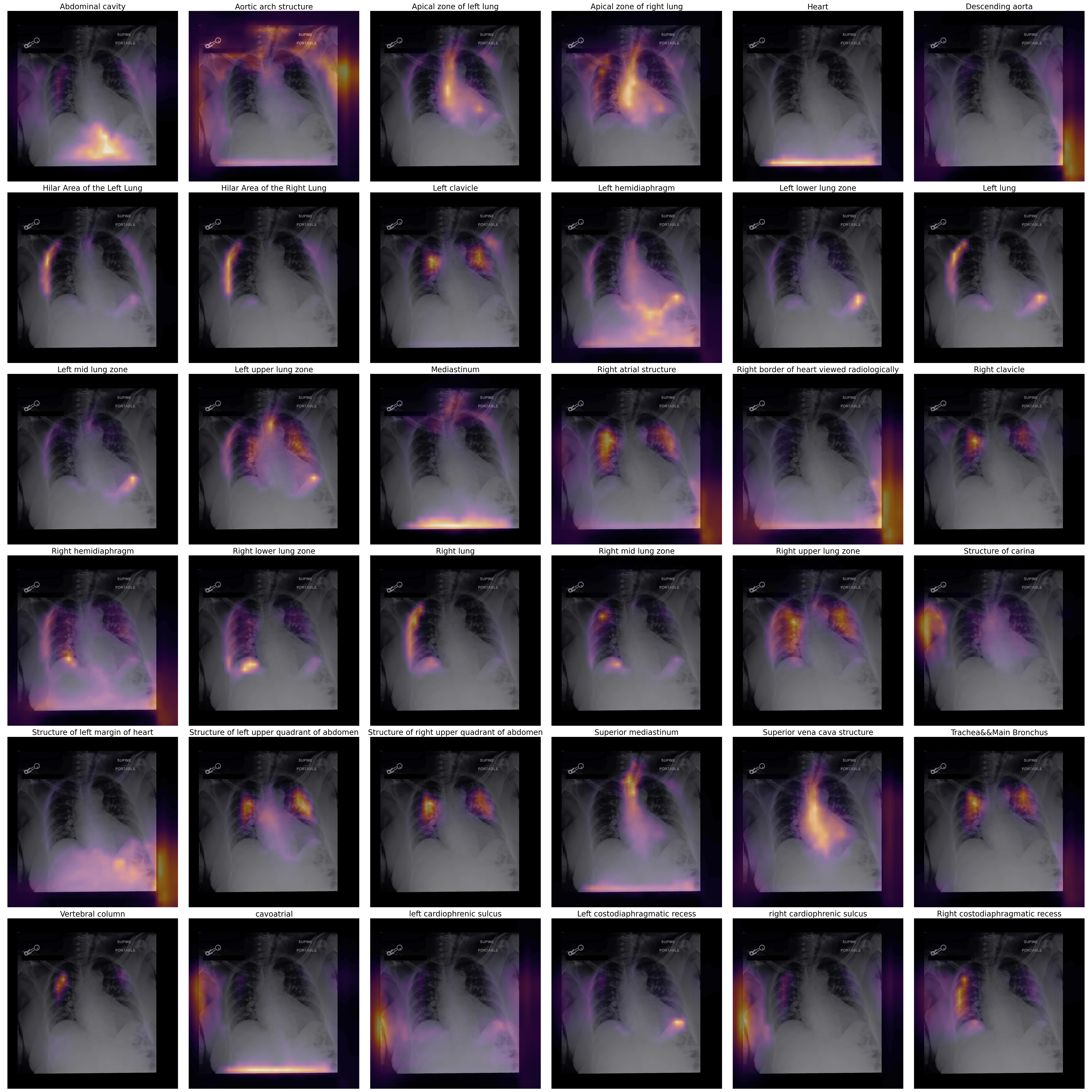}
    \caption{Cross-attention between anatomical object tokens and image embeddings in the feature extraction module $\mathcal{M}$. Subfigure titles indicate the corresponding anatomical object names. Best viewed when zoomed in.}
    \label{fig:attn1}
\end{figure}

In this section, we visualize the cross-attention weights learned within the feature extraction module $\mathcal{M}$ using the anatomical object tokens $\mathrm{O}_A$ as queries and image patch embeddings  $I_p$ as keys. Fig. \ref{fig:attn1} shows that the model focuses on the correct region in the image with high accuracy leading to rich anatomical object tokens $\hat{\mathrm{O}}_A$ and overall anatomical understanding in the downstream tasks.

%% file: sec/ablation.tex
\section{Ablations} \label{sec:ablation}

This section contains the detailed results for the ablations conducted for APM and $\mathcal{LM}$. 


\begin{table}[H]
\caption{Ablation results for grounding tasks. Grounded Diagnosis (GD) and Grounded Captioning (GC) results are given as: GD / GC.}
\centering
\setlength{\tabcolsep}{3.5pt}
\renewcommand{\arraystretch}{1.0}
\scriptsize
\adjustbox{width=\columnwidth,center}{
\begin{tabular}{l|
ccc|cc|cc|cc}
\toprule
\textbf{Model} &
\multicolumn{3}{c|}{\textbf{NLG (GD/GC)}} &
\multicolumn{2}{c|}{\textbf{Clinical (GD/GC)}} &
\multicolumn{2}{c|}{\textbf{Phrase Gr.}} &
\multicolumn{2}{c}{\textbf{Anatomy Gr.}} \\
\cmidrule(lr){2-4}\cmidrule(lr){5-6}\cmidrule(lr){7-8}\cmidrule(lr){9-10}
& BERTScore & ROUGE & METEOR & RadGraph-F1 & CheXbert-14-F1 & IoU & mAP & IoU & mAP \\
\midrule

AnatomiX-$\mathrm{I}_p$ & 0.10 / 0.06  & 0.11 / 0.04  & 0.07 / 0.04  & 0.08 / 0.05  & 0.25 / 0.21 & 0.10 & 0.03 & 0.04 & 0.01 \\

AnatomiX-$\hat{\mathrm{O}}_A$ & 0.42 / 0.17  & 0.38 / 0.12  & 0.26 / 0.06  & 0.35 / 0.08 & 0.42 / 0.23 & 0.24 & 0.16 & 0.36 & 0.27 \\

AnatomiX-$\hat{S}_t$ & 0.19 / 0.25 & 0.17 / 0.23 & 0.16 / 0.18 & 0.23 / 0.22 & 0.28 / 0.24  & 0.11 & 0.05 & 0.06 & 0.02 \\

AnatomiX-$\hat{y}_{box}$ & 0.31 / 0.23 & 0.34 / 0.21 & 0.25 / 0.13 & 0.37 / 0.25 & 0.40 / 0.39 & 0.17 & 0.12 & 0.46 & 0.37  \\

AnatomiX-$\hat{S}_t$-$\hat{y}_{box}$ & 0.49 / 0.45 & 0.52 / 0.36 & 0.34 / 0.26 & 0.51 / 0.34 & 0.49 / 0.61 & 0.26 & 0.17 & 0.47 & 0.35\\

AnatomiX-$\hat{O}_A$-$\hat{y}_{box}$ & 0.52 / 0.40 & \underline{0.58} / 0.33 & \underline{0.37} / 0.28 & 0.62 / 0.37 & 0.50 / \underline{0.67} & 0.36  & 0.24 & \underline{0.61} & \underline{0.53} \\

AnatomiX-$\hat{O}_A$-$\hat{S}_t$ & \underline{0.56} / \underline{0.48} & 0.54 / \underline{0.45} & 0.36 / \underline{0.31} & \underline{0.62} / \underline{0.48} & \underline{0.51} / 0.66 & \underline{0.42} & \underline{0.31} & 0.58 & 0.49 \\

\textbf{AnatomiX} & \textbf{0.63} / \textbf{0.65} & \textbf{0.60} / \textbf{0.56} & \textbf{0.42} / \textbf{0.48} & \textbf{0.58} / \textbf{0.50} & \textbf{0.54} / \textbf{0.78} & \textbf{0.46} & \textbf{0.35} & \textbf{0.73} & \textbf{0.66} \\

\bottomrule
\end{tabular}
}
\label{tab:ablations_grounding_tasks}
\end{table}


\begin{table}[H]
\caption{Ablations results for report generation task grouped by NLG and Clinical metrics.}
\centering
\setlength{\tabcolsep}{5pt}
\renewcommand{\arraystretch}{1.0}

\begin{adjustbox}{max width=\linewidth}
\begin{tabular}{l|ccc|cc}
\toprule
\textbf{Model} &
\multicolumn{3}{c|}{\textbf{NLG Metrics}} &
\multicolumn{2}{c}{\textbf{Clinical Metrics}} \\
\cmidrule(lr){2-4} \cmidrule(lr){5-6}
& ROUGE & BERTScore & METEOR
& RadGraph & CheXbert-14 F1 \\
\midrule

AnatomiX-$\mathrm{I}_p$       & 0.15 & 0.18 & 0.09 & 0.15 & 0.24 \\

AnatomiX-$\hat{\mathrm{O}}_A$ & 0.14 & 0.18 & 0.09 & 0.13 & 0.22 \\

AnatomiX-$\hat{S}_t$ & 0.32 &  0.27 & 0.13 & 0.19 & 0.30 \\ 

AnatomiX-$\hat{y}_{box}$ & 0.13 & 0.15 & 0.10 & 0.11 & 0.21 \\ 

AnatomiX-$\hat{S}_t$-$\hat{y}_{box}$ & 0.27 & 0.24 & 0.10 & 0.15 & 0.23 \\ 

AnatomiX-$\hat{O}_A$-$\hat{y}_{box}$ & 0.15 & 0.21 & 0.11 & 0.14 & 0.22 \\ 

AnatomiX-$\hat{O}_A$-$\hat{S}_t$ & \underline{0.46} & \underline{0.33} & \underline{0.19} & \underline{0.23} & \underline{0.39} \\

\textbf{AnatomiX} & \textbf{0.53} & \textbf{0.38} & \textbf{0.21} & \textbf{0.26} & \textbf{0.42} \\

\bottomrule

\end{tabular}
\end{adjustbox}

\label{tab:ablations_result_report_gen}
\end{table}



\subsection{LLM}

The naming convention for $\mathcal{LM}$'s ablations is as follows: (1) AnatomiX-$\mathrm{I}_p$ uses only image embeddings $\mathrm{I}_p$; (2) AnatomiX-$\hat{\mathrm{O}}_A$ augments $\mathrm{I}_p$ with anatomical tokens $\hat{\mathrm{O}}_A$; (3) AnatomiX-$\hat{S}_t$ combines retrieved sentences $\hat{S}_t$ with $\mathrm{I}_p$; (4) AnatomiX-$\hat{y}_{box}$ integrates predicted bounding boxes $\hat{y}_{box}$ with $\mathrm{I}_p$; (5) AnatomiX-$\hat{S}_t$-$\hat{y}_{box}$ uses $\hat{S}_t$, $\hat{y}_{box}$, and $\mathrm{I}_p$; (6) AnatomiX-$\hat{O}_A$-$\hat{y}_{box}$ combines $\hat{O}_A$, $\hat{y}_{box}$, and $\mathrm{I}_p$; and (7) AnatomiX-$\hat{O}_A$-$\hat{S}_t$ incorporates $\hat{O}_A$, $\hat{S}_t$, and $\mathrm{I}_p$.

Results on grounding tasks (Table~\ref{tab:ablations_grounding_tasks}) indicate substantial performance degradation when anatomical tokens, bounding boxes, and retrieved sentences are removed. Incorporating predicted boxes $\hat{y}_{box}$ markedly improves anatomy grounding and further benefits other tasks when combined with $\hat{S}_t$ and $\hat{O}_A$. Anatomical tokens $\hat{O}_A$ particularly enhance phrase grounding and anatomy grounding, but contribute less to grounded captioning, which requires detailed descriptions; in this setting, retrieved sentences $\hat{S}_t$ provide clear gains. Consistently, report generation results (Table~\ref{tab:ablations_result_report_gen}) show that adding $\hat{S}_t$ yields the largest improvement, underscoring its importance for descriptive generation. Similar patterns are observed in VQA and image understanding (Tables~\ref{tab:ablations_results_image_understanding} and \ref{tab:ablatiosn_vqa_results}), where $\hat{O}_A$ supports spatial reasoning, while $\hat{S}_t$ primarily benefits linguistically intensive tasks. Overall, combining all components achieves the best performance across tasks, particularly demonstrating strong anatomical understanding.

\begin{table}[H]
\caption{Performance on image classification and abnormality detection tasks.}
\centering
\setlength{\tabcolsep}{10pt}
\renewcommand{\arraystretch}{1.0}

\begin{tabular}{l|c|cc}
\toprule
\textbf{Model} &
\multicolumn{1}{c|}{\textbf{Classification}} &
\multicolumn{2}{c}{\textbf{Detection}} \\
\cmidrule(lr){2-2}\cmidrule(lr){3-4} & CheXbert-14 F1
& IoU & mAP \\
\midrule

AnatomiX-$\hat{\mathrm{O}}_A$   & 0.77 & 0.23 & 0.14 \\

AnatomiX-$\hat{S}_t$  & 0.81 & 0.11 & 0.05 \\

AnatomiX-$\hat{y}_{box}$ & 0.78 & 0.18 & 0.10 \\

AnatomiX-$\hat{S}_t$-$\hat{y}_{box}$  & 0.83 & 0.21 & 0.13 \\

AnatomiX-$\hat{O}_A$-$\hat{y}_{box}$ & 0.79 & 0.28 & 0.19 \\

AnatomiX-$\hat{O}_A$-$\hat{S}_t$ & \underline{0.84} & \underline{0.29} & \underline{0.19} \\

AnatomiX-$\mathrm{I}_p$         & 0.75 & 0.08 & 0.06 \\

\textbf{AnatomiX} & \textbf{0.85} & \textbf{0.31} & \textbf{0.20} \\

\bottomrule
\end{tabular}

\label{tab:ablations_results_image_understanding}
\end{table}

\begin{table}[H]
\caption{Open and close ended VQA task performance.}
\centering
\setlength{\tabcolsep}{10pt}  
\renewcommand{\arraystretch}{1.0} 

\begin{tabular}{l|cc|cc}
\toprule
\multirow{2}{*}{\textbf{Model}} &
\multicolumn{2}{c|}{\textbf{Open-Ended VQA}} &
\multicolumn{2}{c}{\textbf{Close-Ended VQA}} \\
\cmidrule(lr){2-3} \cmidrule(lr){4-5}
& BERTScore & CheXbert-14 F1
& BERTScore & CheXbert-14 F1 \\
\midrule

AnatomiX-$\hat{\mathrm{O}}_A$    & 0.70 & 0.74 & 0.73 & 0.91 \\

AnatomiX-$\hat{S}_t$  & 0.81 & 0.84 & 0.81 & 0.94 \\

AnatomiX-$\hat{y}_{box}$ & 0.67 & 0.72 & 0.76 & 0.90 \\

AnatomiX-$\hat{S}_t$-$\hat{y}_{box}$  & 0.78 & 0.79 & 0.83 & 0.95 \\

AnatomiX-$\hat{O}_A$-$\hat{y}_{box}$ & 0.73 & 0.75 & 0.79 & 0.85 \\

AnatomiX-$\hat{O}_A$-$\hat{S}_t$ & \underline{0.82} & \underline{0.85} & \underline{0.86} & \textbf{0.95} \\

AnatomiX-$\mathrm{I}_p$          & 0.68 & 0.72 & 0.72 & 0.87 \\

\textbf{AnatomiX} & \textbf{0.86} & \textbf{0.86} & \textbf{0.89} & \underline{0.95} \\

\bottomrule
\end{tabular}

\label{tab:ablatiosn_vqa_results}
\end{table}


\subsection{APM}
In addition to the ablations conducted for $\mathcal{LM}$, we study the contribution of different architectural components in APM. First, we replace the image encoder $\mathcal{E}$ with a pretrained DINOv3 model, denoted as AnatomiX-Dino. Second, to evaluate the role of the feature extraction module $\mathcal{M}$, we remove it and perform contrastive alignment directly on the decoder output $\mathcal{D}$, resulting in AnatomiX-wo-$\mathcal{M}$. Third, in APM-CLIP, we replace the proposed contrastive self-similarity loss with the standard CLIP loss. We evaluate all variants on object detection (bounding box prediction $y_{box}$) and sentence retrieval. For detection, we report Intersection over Union (IoU). For retrieval, we compare the retrieved sentence $\hat{S}_t$ with ground-truth text using CheXbert-14-F1, RadGraph-F1, and METEOR.

As shown in Table~\ref{tab:ablation_APM}, DINOv3 achieves competitive IoU but underperforms on retrieval metrics, indicating weaker cross-modal alignment despite strong visual representations. Removing $\mathcal{M}$ results in a consistent performance drop across tasks, suggesting that decoupling bounding box prediction from textual feature alignment facilitates more effective learning and improves both anatomical localization and sentence retrieval. Finally, replacing the proposed soft self-similarity loss with the standard CLIP loss degrades performance, highlighting the importance of the tailored contrastive objective in APM. Furthermore, anatomy-wise breakdown of retrieval results in given in Tables \ref{tab:mid_with_laterality_anatomy_structure_results}-\ref{tab:mid_anatomy_structure_results} (d-g).

\begin{table}[H]
\caption{Ablation experiments for APM discussed in S5.}
\centering
\setlength{\tabcolsep}{5pt}
\renewcommand{\arraystretch}{1.0}

\begin{tabular}{l|c|ccc}
\toprule
& \multicolumn{1}{c|}{\textbf{$y_{box}$ Metrics}} & \multicolumn{3}{c}{\textbf{$\hat{S}_t$ Metrics}} \\
\cmidrule(lr){2-2}\cmidrule(lr){3-5}
\textbf{Model} & IoU & CheXbert-14 F1 & RadGraph F1 & METEOR \\
\midrule

APM-wo-$\mathcal{M}$ & 0.781 & 0.627 & \underline{0.689} & \underline{0.710} \\

APM-Dino & \underline{0.792} & 0.611 & 0.685 & 0.704 \\

APM-CLIP & 0.775 & \textbf{0.640} & 0.679 & 0.692 \\

APM & \textbf{0.812} & \underline{0.634} & \textbf{0.709} & \textbf{0.727} \\

\bottomrule
\end{tabular}

\label{tab:ablation_APM}
\end{table}

\setlength{\tabcolsep}{3.5pt}

%% file: main.bib
@String(ICLR = {Int. Conf. Learn. Represent.})

@String(ICLR  = {ICLR})

@inproceedings{carion2020end,
  title={End-to-end object detection with transformers},
  author={Carion, Nicolas and Massa, Francisco and Synnaeve, Gabriel and Usunier, Nicolas and Kirillov, Alexander and Zagoruyko, Sergey},
  booktitle={European conference on computer vision},
  pages={213--229},
  year={2020},
  organization={Springer}
}

@article{sellergren2025medgemma,
  title={Medgemma technical report},
  author={Sellergren, Andrew and Kazemzadeh, Sahar and Jaroensri, Tiam and Kiraly, Atilla and Traverse, Madeleine and Kohlberger, Timo and Xu, Shawn and Jamil, Fayaz and Hughes, C{\'\i}an and Lau, Charles and others},
  journal={arXiv preprint arXiv:2507.05201},
  year={2025}
}

@article{hu2022lora,
  title={Lora: Low-rank adaptation of large language models.},
  author={Hu, Edward J and Shen, Yelong and Wallis, Phillip and Allen-Zhu, Zeyuan and Li, Yuanzhi and Wang, Shean and Wang, Lu and Chen, Weizhu and others},
  journal={ICLR},
  volume={1},
  number={2},
  pages={3},
  year={2022}
}

@inproceedings{radford2021learning,
  title={Learning transferable visual models from natural language supervision},
  author={Radford, Alec and Kim, Jong Wook and Hallacy, Chris and Ramesh, Aditya and Goh, Gabriel and Agarwal, Sandhini and Sastry, Girish and Askell, Amanda and Mishkin, Pamela and Clark, Jack and others},
  booktitle={International conference on machine learning},
  pages={8748--8763},
  year={2021},
  organization={PmLR}
}

@article{wu2021chest,
  title={Chest imagenome dataset for clinical reasoning},
  author={Wu, Joy T and Agu, Nkechinyere N and Lourentzou, Ismini and Sharma, Arjun and Paguio, Joseph A and Yao, Jasper S and Dee, Edward C and Mitchell, William and Kashyap, Satyananda and Giovannini, Andrea and others},
  journal={arXiv preprint arXiv:2108.00316},
  year={2021}
}

@article{johnson2019mimic,
  title={MIMIC-CXR, a de-identified publicly available database of chest radiographs with free-text reports},
  author={Johnson, Alistair EW and Pollard, Tom J and Berkowitz, Seth J and Greenbaum, Nathaniel R and Lungren, Matthew P and Deng, Chih-ying and Mark, Roger G and Horng, Steven},
  journal={Scientific data},
  volume={6},
  number={1},
  pages={317},
  year={2019},
  publisher={Nature Publishing Group UK London}
}

@article{nguyen2022vindr,
  title={VinDr-CXR: An open dataset of chest X-rays with radiologist’s annotations},
  author={Nguyen, Ha Q and Lam, Khanh and Le, Linh T and Pham, Hieu H and Tran, Dat Q and Nguyen, Dung B and Le, Dung D and Pham, Chi M and Tong, Hang TT and Dinh, Diep H and others},
  journal={Scientific Data},
  volume={9},
  number={1},
  pages={429},
  year={2022},
  publisher={Nature Publishing Group UK London}
}

@inproceedings{boecking2022making,
  title={Making the most of text semantics to improve biomedical vision--language processing},
  author={Boecking, Benedikt and Usuyama, Naoto and Bannur, Shruthi and Castro, Daniel C and Schwaighofer, Anton and Hyland, Stephanie and Wetscherek, Maria and Naumann, Tristan and Nori, Aditya and Alvarez-Valle, Javier and others},
  booktitle={European conference on computer vision},
  pages={1--21},
  year={2022},
  organization={Springer}
}

@inproceedings{liu2021slake,
  title={Slake: A semantically-labeled knowledge-enhanced dataset for medical visual question answering},
  author={Liu, Bo and Zhan, Li-Ming and Xu, Li and Ma, Lin and Yang, Yan and Wu, Xiao-Ming},
  booktitle={2021 IEEE 18th international symposium on biomedical imaging (ISBI)},
  pages={1650--1654},
  year={2021},
  organization={IEEE}
}

@misc{bae2024mimic,
  title={Mimic-ext-mimic-cxr-vqa: A complex, diverse, and large-scale visual question answering dataset for chest x-ray images},
  author={Bae, Seongsu and Kyung, Daeun and Ryu, Jaehee and Cho, Eunbyeol and Lee, Gyubok and Kweon, Sunjun and Oh, Jungwoo and Ji, Lei and Chang, Eric and Kim, Tackeun and others},
  year={2024},
  publisher={PhysioNet}
}

@article{pellegriniradialog,
  title={RaDialog Instruct Dataset},
  author={Pellegrini, Chantal and {\"O}zsoy, Ege and Busam, Benjamin and Navab, Nassir and Keicher, Matthias}
}

@article{de2025padchest,
  title={Padchest-gr: A bilingual chest X-ray dataset for grounded radiology report generation},
  author={de Castro, Daniel Coelho and Bustos, Aurelia and Bannur, Shruthi and Hyland, Stephanie L and Bouzid, Kenza and Wetscherek, Maria Teodora and S{\'a}nchez-Valverde, Maria Dolores and Jaques-P{\'e}rez, Lara and P{\'e}rez-Rodr{\'\i}guez, Lourdes and Takeda, Kenji and others},
  journal={NEJM AI},
  volume={2},
  number={7},
  pages={AIdbp2401120},
  year={2025},
  publisher={Massachusetts Medical Society}
}

@article{xiao2024comprehensive,
  title={A comprehensive survey of large language models and multimodal large language models in medicine},
  author={Xiao, Hanguang and Zhou, Feizhong and Liu, Xingyue and Liu, Tianqi and Li, Zhipeng and Liu, Xin and Huang, Xiaoxuan},
  journal={Information Fusion},
  pages={102888},
  year={2024},
  publisher={Elsevier}
}

@article{liu2023visual,
  title={Visual instruction tuning},
  author={Liu, Haotian and Li, Chunyuan and Wu, Qingyang and Lee, Yong Jae},
  journal={Advances in neural information processing systems},
  volume={36},
  pages={34892--34916},
  year={2023}
}

@inproceedings{wolf2025your,
  title={Your other Left! Vision-Language Models Fail to Identify Relative Positions in Medical Images},
  author={Wolf, Daniel and Hillenhagen, Heiko and Taskin, Billurvan and B{\"a}uerle, Alex and Beer, Meinrad and G{\"o}tz, Michael and Ropinski, Timo},
  booktitle={International Conference on Medical Image Computing and Computer-Assisted Intervention},
  pages={691--701},
  year={2025},
  organization={Springer}
}

@article{anisuzzaman2024fine,
  title={Fine-tuning llms for specialized use cases},
  author={Anisuzzaman, DM and Malins, Jeffrey G and Friedman, Paul A and Attia, Zachi I},
  journal={Mayo Clinic Proceedings: Digital Health},
  year={2024},
  publisher={Elsevier}
}

@article{liu2025can,
  title={Can Multimodal Large Language Models Understand Spatial Relations?},
  author={Liu, Jingping and Liu, Ziyan and Cen, Zhedong and Zhou, Yan and Zou, Yinan and Zhang, Weiyan and Jiang, Haiyun and Ruan, Tong},
  journal={arXiv preprint arXiv:2505.19015},
  year={2025}
}

@article{szot2024grounding,
  title={Grounding multimodal large language models in actions},
  author={Szot, Andrew and Mazoure, Bogdan and Agrawal, Harsh and Hjelm, R Devon and Kira, Zsolt and Toshev, Alexander},
  journal={Advances in Neural Information Processing Systems},
  volume={37},
  pages={20198--20224},
  year={2024}
}

@article{peng2023kosmos,
  title={Kosmos-2: Grounding multimodal large language models to the world},
  author={Peng, Zhiliang and Wang, Wenhui and Dong, Li and Hao, Yaru and Huang, Shaohan and Ma, Shuming and Wei, Furu},
  journal={arXiv preprint arXiv:2306.14824},
  year={2023}
}

@article{luo2024vividmed,
  title={Vividmed: Vision language model with versatile visual grounding for medicine},
  author={Luo, Lingxiao and Tang, Bingda and Chen, Xuanzhong and Han, Rong and Chen, Ting},
  journal={arXiv preprint arXiv:2410.12694},
  year={2024}
}

@article{chen2024chexagent,
  title={Chexagent: Towards a foundation model for chest x-ray interpretation},
  author={Chen, Zhihong and Varma, Maya and Delbrouck, Jean-Benoit and Paschali, Magdalini and Blankemeier, Louis and Van Veen, Dave and Valanarasu, Jeya Maria Jose and Youssef, Alaa and Cohen, Joseph Paul and Reis, Eduardo Pontes and others},
  journal={arXiv preprint arXiv:2401.12208},
  year={2024}
}

@article{deperrois2025radvlm,
  title={RadVLM: A multitask conversational vision-language model for radiology},
  author={Deperrois, Nicolas and Matsuo, Hidetoshi and Ruip{\'e}rez-Campillo, Samuel and Vandenhirtz, Moritz and Laguna, Sonia and Ryser, Alain and Fujimoto, Koji and Nishio, Mizuho and Sutter, Thomas M and Vogt, Julia E and others},
  journal={arXiv preprint arXiv:2502.03333},
  year={2025}
}

@article{bannur2024maira,
  title={Maira-2: Grounded radiology report generation},
  author={Bannur, Shruthi and Bouzid, Kenza and Castro, Daniel C and Schwaighofer, Anton and Thieme, Anja and Bond-Taylor, Sam and Ilse, Maximilian and P{\'e}rez-Garc{\'\i}a, Fernando and Salvatelli, Valentina and Sharma, Harshita and others},
  journal={arXiv preprint arXiv:2406.04449},
  year={2024}
}

@inproceedings{ma2024groma,
  title={Groma: Localized visual tokenization for grounding multimodal large language models},
  author={Ma, Chuofan and Jiang, Yi and Wu, Jiannan and Yuan, Zehuan and Qi, Xiaojuan},
  booktitle={European Conference on Computer Vision},
  pages={417--435},
  year={2024},
  organization={Springer}
}

@article{loshchilov2017decoupled,
  title={Decoupled weight decay regularization},
  author={Loshchilov, Ilya and Hutter, Frank},
  journal={arXiv preprint arXiv:1711.05101},
  year={2017}
}

@article{smit2020chexbert,
  title={CheXbert: combining automatic labelers and expert annotations for accurate radiology report labeling using BERT},
  author={Smit, Akshay and Jain, Saahil and Rajpurkar, Pranav and Pareek, Anuj and Ng, Andrew Y and Lungren, Matthew P},
  journal={arXiv preprint arXiv:2004.09167},
  year={2020}
}

@inproceedings{lin2004rouge,
  title={Rouge: A package for automatic evaluation of summaries},
  author={Lin, Chin-Yew},
  booktitle={Text summarization branches out},
  pages={74--81},
  year={2004}
}

@inproceedings{banerjee2005meteor,
  title={METEOR: An automatic metric for MT evaluation with improved correlation with human judgments},
  author={Banerjee, Satanjeev and Lavie, Alon},
  booktitle={Proceedings of the acl workshop on intrinsic and extrinsic evaluation measures for machine translation and/or summarization},
  pages={65--72},
  year={2005}
}

@article{delbrouck2022improving,
  title={Improving the factual correctness of radiology report generation with semantic rewards},
  author={Delbrouck, Jean-Benoit and Chambon, Pierre and Bluethgen, Christian and Tsai, Emily and Almusa, Omar and Langlotz, Curtis P},
  journal={arXiv preprint arXiv:2210.12186},
  year={2022}
}

@article{zhang2019bertscore,
  title={Bertscore: Evaluating text generation with bert},
  author={Zhang, Tianyi and Kishore, Varsha and Wu, Felix and Weinberger, Kilian Q and Artzi, Yoav},
  journal={arXiv preprint arXiv:1904.09675},
  year={2019}
}

@article{gu2021domain,
  title={Domain-specific language model pretraining for biomedical natural language processing},
  author={Gu, Yu and Tinn, Robert and Cheng, Hao and Lucas, Michael and Usuyama, Naoto and Liu, Xiaodong and Naumann, Tristan and Gao, Jianfeng and Poon, Hoifung},
  journal={ACM Transactions on Computing for Healthcare (HEALTH)},
  volume={3},
  number={1},
  pages={1--23},
  year={2021},
  publisher={ACM New York, NY}
}

@article{li2023llava,
  title={Llava-med: Training a large language-and-vision assistant for biomedicine in one day},
  author={Li, Chunyuan and Wong, Cliff and Zhang, Sheng and Usuyama, Naoto and Liu, Haotian and Yang, Jianwei and Naumann, Tristan and Poon, Hoifung and Gao, Jianfeng},
  journal={Advances in Neural Information Processing Systems},
  volume={36},
  pages={28541--28564},
  year={2023}
}

@inproceedings{kirillov2023segment,
  title={Segment anything},
  author={Kirillov, Alexander and Mintun, Eric and Ravi, Nikhila and Mao, Hanzi and Rolland, Chloe and Gustafson, Laura and Xiao, Tete and Whitehead, Spencer and Berg, Alexander C and Lo, Wan-Yen and others},
  booktitle={Proceedings of the IEEE/CVF international conference on computer vision},
  pages={4015--4026},
  year={2023}
}

@article{zou2024medrg,
  title={Medrg: Medical report grounding with multi-modal large language model},
  author={Zou, Ke and Bai, Yang and Chen, Zhihao and Zhou, Yang and Chen, Yidi and Ren, Kai and Wang, Meng and Yuan, Xuedong and Shen, Xiaojing and Fu, Huazhu},
  journal={arXiv preprint arXiv:2404.06798},
  year={2024}
}

@inproceedings{muller2024chex,
  title={ChEX: Interactive localization and region description in chest X-rays},
  author={M{\"u}ller, Philip and Kaissis, Georgios and Rueckert, Daniel},
  booktitle={European Conference on Computer Vision},
  pages={92--111},
  year={2024},
  organization={Springer}
}

@article{li2025aor,
  title={AOR: Anatomical Ontology-Guided Reasoning for Medical Large Multimodal Model in Chest X-Ray Interpretation},
  author={Li, Qingqiu and Cui, Zihang and Bae, Seongsu and Xu, Jilan and Yuan, Runtian and Zhang, Yuejie and Feng, Rui and Shen, Quanli and Zhang, Xiaobo and He, Junjun and others},
  journal={arXiv preprint arXiv:2505.02830},
  year={2025}
}

@inproceedings{jiang2023cross,
  title={Cross-modal implicit relation reasoning and aligning for text-to-image person retrieval},
  author={Jiang, Ding and Ye, Mang},
  booktitle={Proceedings of the IEEE/CVF conference on computer vision and pattern recognition},
  pages={2787--2797},
  year={2023}
}

@inproceedings{kovatch2020optimizing,
  author    = {Kovatch, P. and Gai, L. and Cho, H. M. and Fluder, E. and Jiang, D.},
  title     = {Optimizing High-Performance Computing Systems for Biomedical Workloads},
  booktitle = {Proceedings of the IEEE International Symposium on Parallel and Distributed Processing Workshops (IPDPSW) PhD Forum},
  year      = {2020},
  month     = {May},
  pages     = {183--192},
  doi       = {10.1109/IPDPSW50202.2020.00040},
  note      = {Epub 2020-07-28},
  pmid      = {33088611},
  pmcid     = {PMC7575271}
}
